\documentclass[final,5p,times,twocolumn,number]{elsarticle}
\usepackage{xurl}
\usepackage{hyperref}
\usepackage{amssymb}
\usepackage{lipsum}
\usepackage{amsmath}
\usepackage{booktabs}
\usepackage{siunitx}
\usepackage{blindtext}
\usepackage{titlesec}
\usepackage{xcolor}
\usepackage{hyperref}
\hypersetup{colorlinks=true, citecolor=red, linkcolor=blue, filecolor=blue}
\usepackage[nameinlink]{cleveref}
\crefname{figure}{Fig.}{Figs.}
\usepackage{bm}
\usepackage{stfloats}
\usepackage{caption}
\usepackage{graphicx} 
\usepackage{enumitem}
\usepackage{makecell}
\usepackage{multirow}
\biboptions{sort&compress}  

\usepackage[switch]{lineno}
\makeatletter
\def\ps@pprintTitle{%
 \let\@oddhead\@empty
 \let\@evenhead\@empty
 \def\@oddfoot{}%
 \let\@evenfoot\@oddfoot}
\makeatother

\begin{document}

\begin{frontmatter}

\title{Contact-Rich Robotic Manipulation in Construction via Zero-Shot Learning: A Diffusion Policy-Guided Adaptive Control} 


\author[add1]{Roman Ibrahimov\corref{cor1}}
\ead{roman.ibrahimov@princeton.edu}

\author[add1]{Salma Mozaffari\corref{cor1}}
\ead{salma.mozaffari@princeton.edu}

\author[add1]{Arash Adel\corref{cor2}}
\ead{arash.adel@princeton.edu}

\address[add1]{Princeton University, Princeton, NJ 08544, USA}
           
\cortext[cor1]{Equal contribution.}           
\cortext[cor2]{Corresponding author.}


\begin{abstract}

Construction robotics and automation offer promising means of improving productivity, alleviating workforce shortages, and reducing workers' exposure to physically demanding tasks. However, reliable contact-rich robotic assembly remains challenging under tight tolerances, fabrication inaccuracies, and uncertain contact dynamics. To address this challenge, we present a framework coupling diffusion policies trained on simulation-generated pose and force/torque data with an $\mathcal{L}_1$-inspired adaptive controller that corrects policy-predicted actions online to compensate for unmodeled contact dynamics. We benchmark the framework against baselines in timber joinery, pipe fitting, and sequential full-scale truss assembly. It achieves 100\% success on single-task assemblies and 90--100\% success across sequential truss assembly subtasks, with lower, more stable contact forces than the baselines. By enabling zero-shot sim-to-real transfer for force-aware contact-rich assembly, the framework reduces costly, labor-intensive real-world data collection for policy training and advances scalable, robust automation of multistage assembly, motivating extension to broader contact-rich manipulation tasks in construction.

\end{abstract}



\end{frontmatter}

\begin{figure*}[ht]
    \centering
    \includegraphics[width=\textwidth]{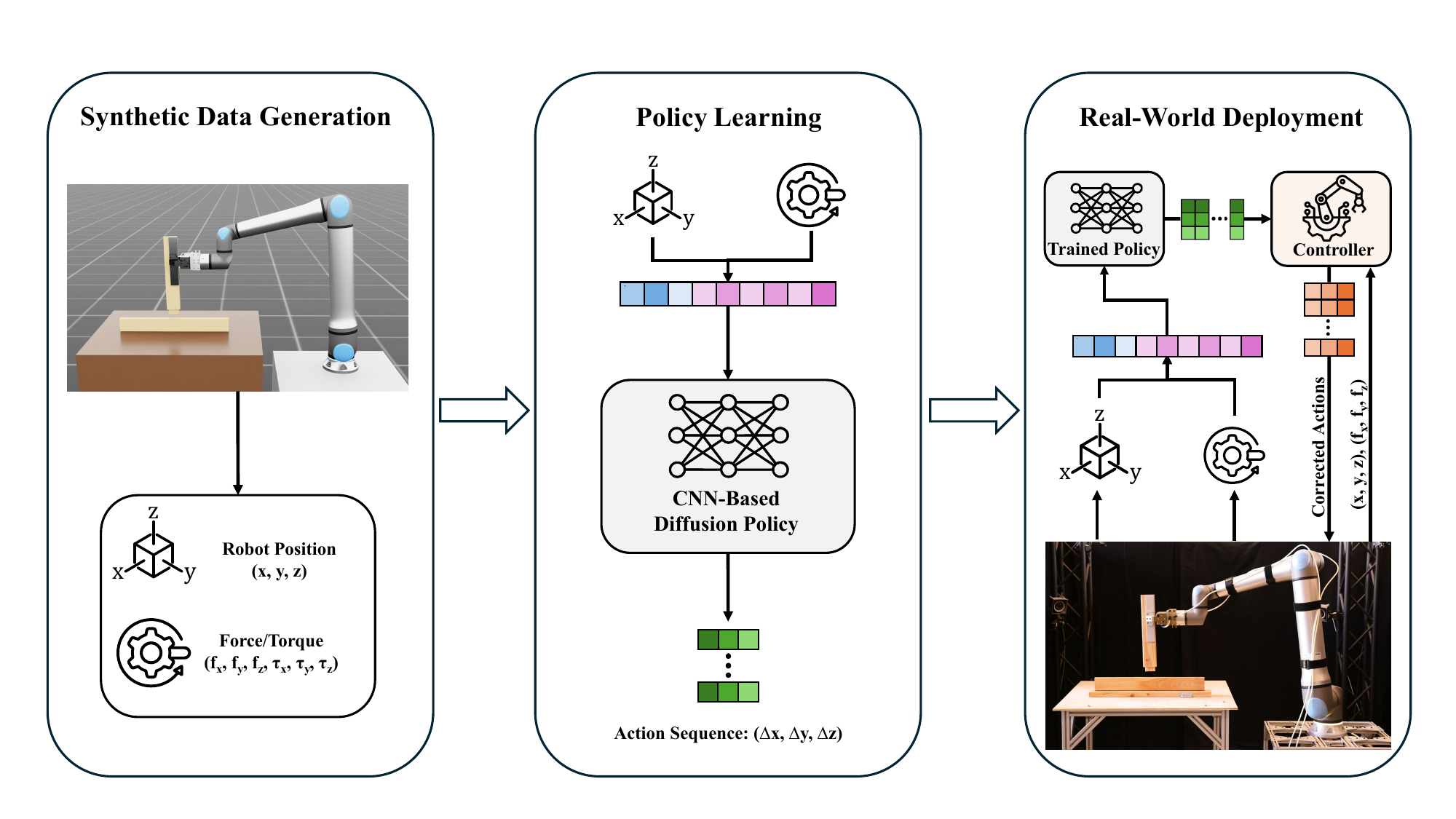}
    \caption{Methods overview: we generate synthetic data in simulation using the robot position at the flange center and force/torque feedback at the robot flange (left) to train a CNN-based diffusion policy (center), then, we develop an adaptive controller to correct the policy-predicted actions ($\Delta x, \Delta y, \Delta z$) of the trained policy, facilitating a zero-shot policy transfer and improving performance in real-world deployment (right).}
    \label{fig:overview}
\end{figure*}

\section{Introduction}\label{introduction}

The construction industry accounts for a substantial share of global economic activity~\cite{mckinsey2024}. Despite its economic importance, the industry continues to experience low productivity growth relative to sectors such as manufacturing~\cite{Bock2015, delgado2019}. In addition, the industry faces substantial workforce shortages. For instance, the U.S. construction industry is projected to need to attract 456,000 additional workers in 2027 to meet industry demand~\cite{ABC2026}. Construction work also involves physically demanding tasks, including lifting and positioning heavy components, that expose workers to elevated risks of occupational injuries and work-related musculoskeletal disorders~\cite{Laukkanen1999, Abdelhamid2002, Arndt2005, Hwang2017}. Construction robotics and automation have therefore been proposed as promising means of improving productivity, alleviating workforce constraints, and reducing workers' exposure to physically demanding tasks~\cite{Bock2015, liang2020, Wang2023, adel2024, chen2025}. 

Realizing these potential benefits, however, depends in part on robotic systems that can reliably manipulate full-scale building components by grasping, transporting, orienting, placing, and joining them while handling physical interactions with materials and the evolving as-built structure~\cite{giftthaler2017, chen2025, mozaffari2026}. Within this broader domain of manipulation, assembly is especially critical because many building systems, including walls, floors, building services, and structural frames, are constructed by joining discrete components into larger systems. Robotic assembly therefore offers a path toward automating labor-intensive and physically demanding construction tasks that require substantial trade expertise. Many demonstrations of robotic assembly in construction, as illustrated by nonstandard brickwork and timber structures~\cite{bonwetsch2016, willmann2016, apolinarska2018, adel2020, Adel2022}, can be effective when calibration, planned tolerances and subsequent joining techniques can compensate for positioning errors and accumulated assembly tolerances. However, for assembly processes involving tight-fitting connections, executing preplanned trajectories without feedback can be unreliable or result in assembly failure because fabrication inaccuracies and accumulated tolerances can create considerable deviations between the as-planned and as-built geometries~\cite{ruan2023, adel2024}.

To address this challenge, prior work in timber assembly has incorporated perception-based feedback and adaptive fabrication to update subsequent fabrication steps and placement poses (i.e., positions and orientations) in response to deviations between as-planned and as-built states~\cite{ruan2023, adel2024, cote2024, adel2026, gao2026}. While these projects have been successful for demonstrating feedback-driven assembly, they mainly rely on planar face-to-face connections or butt joins. These approaches do not directly address contact-rich construction assembly involving high friction and tight clearances, such as mating structural components (e.g., framing using timber joinery) or assembling building service components (e.g., pipe fitting). In such tasks, tight clearances, fabrication inaccuracies, and unmodeled contact dynamics can result in unreliable execution or assembly failure, while excessive contact forces can damage components. Successful execution therefore depends not only on accurate positioning but also on robust control during contact. Incorporating force feedback is thus crucial for closing the loop and enabling corrective actions in response to positional and contact uncertainty during task execution.

To address this challenge, learning-based methods have been explored for contact-rich assembly in construction~\cite{apolinarska2021, mozaffari2026}. In particular, sensory--motor policy learning methods learn end-to-end mappings from sensory observations, such as camera images and robot state, directly to robot actions~\cite{finn2017, levine2018, kalashnikov2018, brohan2023}. Among these methods, behavior cloning through diffusion policy~\cite{chi2024a} has demonstrated strong performance across diverse robotic manipulation tasks \cite{chi2024b, hou2025, yang2025, huaijiang2025}. Prior research has investigated training diffusion policies for contact-rich assembly in construction using pose and force/torque (F/T) feedback to mate timber joints with sub-millimeter clearance under fabrication uncertainty~\cite{mozaffari2026}. This prior study demonstrated the potential of diffusion-based control policy and the contribution of F/T feedback for precise assembly under uncertain contact dynamics and positional misalignments. However, this prior research also indicated that training the policies required extensive expert demonstrations on a physical robot, making data collection time-intensive and contributing to degradation of the repeatedly assembled workpieces. Repeated physical trials may also increase equipment wear.

Accordingly, the research presented in this paper aims to enable robust, force-aware contact-rich assembly of construction-scale components without requiring physical demonstrations for policy training. At this scale, real-world data collection is costly, time-consuming, and labor-intensive, and unsuccessful trials can generate excessive contact forces that damage workpieces and accelerate equipment wear. We therefore collect all policy-training data in simulation. This approach, however, must contend with discrepancies between the simulated and physical systems in component geometry, material properties, and contact dynamics, which can introduce a significant sim-to-real gap. To address this challenge, we present, evaluate, and benchmark an integrated framework in which a diffusion policy trained on simulation-generated robot position and F/T data predicts actions, while an adaptive controller corrects these actions online to compensate for unmodeled contact dynamics and material imperfections (\cref{fig:overview}). The proposed framework enables zero-shot transfer from simulation-only policy training to force-aware execution of contact-rich construction assembly. 

In the following subsection, we summarize the primary contributions of this paper.

\subsection{Contributions}

\begin{itemize}[itemsep=1pt, topsep=2pt] 

\item We introduce a scalable synthetic data generation pipeline that enables training diffusion policies in simulation and deploying them on a construction-scale physical robot for contact-rich manipulation tasks. The pipeline generates diverse trajectories with robot position and F/T data, significantly reducing reliance on costly, time-consuming, and labor-intensive real-world demonstrations.

\item We develop a control architecture that integrates learned diffusion policies with a controller inspired by $\mathcal{L}_1$ adaptive control~\cite{hovakimyan2010}. The controller corrects policy-predicted actions online to improve robustness to material imperfections and unmodeled contact dynamics during tight-tolerance physical assembly. To the best of our knowledge, this paper represents one of the first developments of this novel control architecture for construction-scale robotic assembly.

\item We experimentally evaluate and demonstrate that combining diffusion policy with the proposed adaptive controller enables zero-shot sim-to-real transfer with improved success rates in simulation and real. In addition to high task success, the framework produces lower and more stable contact forces, leading to safer and more controlled interactions. Furthermore, the framework enables large-scale simulation rollouts, allowing extensive testing and more representative evaluation beyond what is feasible with physical experiments alone.

\item In addition to single assembly tasks, we validate the proposed framework in a full-scale timber prefabrication experiment where separate trained policies are deployed sequentially to assemble a gable-shaped truss through multiple contact-rich insertions. This experiment demonstrates that our proposed method can be integrated into a construction-scale assembly workflow to facilitate consecutive high-precision contact-rich subassembly tasks.
   
\end{itemize}

\section{Related work}\label{sec:related_work}

This section situates the proposed framework within four complementary areas of research relevant to robust, contact-rich robotic assembly in construction. First, we review robotic assembly in construction, tracing the progression from preplanned workflows to systems that use perception and learned policies to address fabrication uncertainty at construction scale. Second, we examine classical model-based approaches, including hybrid force--position, impedance, and variable-impedance control, which provide explicit structures for compliant interaction. Third, we review learning-based methods, with particular attention to multimodal sensory--motor policies and sim-to-real transfer. Fourth, we examine learning-integrated model-based controllers that combine learned motion or compliance with structured force--position, impedance, or admittance control.

\subsection{Robotic assembly in construction}
\label{sec:robotic_assembly_construction}

Many prior studies in construction robotics has focused on assembly processes that rely primarily on accurate positioning and sequencing of elements, often assuming precise calibration with negligible uncertainties as well as simple planar connections. Relevant examples include robotic brickwork realizing differentiated assemblies through repeatable pick-and-place operations~\cite{bonwetsch2015}. Similar studies enable spatial and customized robotic assembly for timber and steel frames~\cite{willmann2016, parascho2017, adel2018, adel2020}. These studies demonstrated the feasibility of customized, full-scale robotic assembly, but most provided limited online correction once contact-induced deviations emerged during execution. Such approaches are well-suited for assembly tasks involving loose tolerances or limited physical interactions, where moderate deviations do not critically affect task success. 

In contrast, the assembly tasks involving contact-rich manipulation, such as timber framing with tight-fitting interlocking joints, introduce contact sensitivity, geometric constraints, and the possibility of misalignment~\cite{helmreich2022,leung2021}. In these tasks, small deviations due to inherent uncertainties such as calibration errors, fabrication inaccuracies, or material imperfections can cause misalignment, jamming, or incomplete mating, making execution based only on precomputed motion unreliable. 

Prior work has addressed fabrication uncertainty through computational models of tolerance propagation and assembly-sequence optimization intended to minimize accumulated positional error~\cite{gandia2022}. While such approaches improve robustness at the planning stage, execution is typically carried out using predefined trajectories with limited real-time adjustments. As a result, there has been a growing shift toward closed-loop methods that leverage sensory feedback. Examples include iterative pose correction to compensate for deviations between the as-planned digital model and the as-built structure during multi-robot timber assembly~\cite{ruan2023, adel2024} and using computer vision for component pose estimation and alignment in rebar tying and wood assembly~\cite{liu2025, cote2024}. While these methods improve planning, localization, or component pose correction, they do not necessarily provide online correction of actions from measured interaction forces during complex contact interactions.

Recent research in construction robotics has explored applying learned policies to contact-rich assembly. Examples include reinforcement learning and diffusion policy learning from pose and force/torque feedback for assembly using timber joinery~\cite{apolinarska2021, mozaffari2026}; vision-based hand gesture control for collecting intuitive demonstrations for imitation learning~\cite{duan2024}; reinforcement learning with domain adaptation using real tactile images for sim-to-real cable-in-duct installation~\cite{duan2025}; policy learning from human demonstration videos for ceiling tile installation evaluated in Gazebo~\cite{liang2020}; and imitation learning for rebar insertion and tying~\cite{sun2026}. 

The data collection methods of these studies differ substantially. Mozaffari et al. use 400 teleoperated demonstrations collected on a physical robot~\cite{mozaffari2026}, whereas Liang et al. use 85 real and 3,000 virtual human-demonstration videos rather than expert demonstrations collected on the robot itself~\cite{liang2020}. Duan et al. use domain adaptation with real tactile images to improve simulation-based reinforcement learning and sim-to-real transfer~\cite{duan2025}. Learning from demonstrations on physical robots can therefore incur substantial data collection cost, while simulation-based approaches reduce that burden but must still address discrepancies between the sim and real. Physical evaluation of trained models and task-specific fine-tuning can also remain costly and potentially hazardous at construction scale because repeated contact can degrade materials and wear equipment. These limitations motivate methods that combine scalable simulation-based training with force-aware action correction during physical execution. 

\subsection{Model-based control}
\label{sec:model_based_control}

Model-based control methods for contact-rich manipulation have a long history in hybrid force--position control, motivated by the need to regulate physical interaction under geometric uncertainty and unmodeled contact dynamics. Early foundational work introduced hybrid force--position control, formalizing the decomposition of task space into position- and force-controlled subspaces to explicitly comply with contact constraints~\cite{raibert1981, mason2007}. This line of research was complemented by impedance control, which framed interaction as the regulation of a desired mechanical impedance rather than precise force tracking, enabling compliant behavior to emerge naturally during contact~\cite{hogan1985}. Contact-rich manipulation tasks often involve distinct phases with different force requirements. Variable impedance control extends fixed-impedance formulations by adapting stiffness and damping across these phases, improving robustness in tasks such as peg-in-hole insertion~\cite{yang2022}.

Nevertheless, classical model-based control formulations alone do not fully resolve the challenges posed by contact-rich manipulation under uncertainty. Classical hybrid and impedance controllers generally require predefined task frames, contact constraints, desired impedances, or switching logic. Their performance can therefore depend on the accuracy of the task representation, the appropriateness of the assumed contact structure, and task-specific controller design or tuning. Also, geometric deviations, frictional variability, and material compliance can vary significantly across instances, limiting the effectiveness of fixed model assumptions. More recent research projects have explored the integration of learning-based control within model-based control architectures to address these challenges. These methods are discussed in \ref{sec:Learning_model_based_control}.

\subsection{Learning-based control}
\label{sec:learning_based_control}

Learning-based control methods have demonstrated strong performance in implicitly capturing contact dynamics that are difficult to model analytically. Early work showed that end-to-end visuomotor policies trained from demonstrations can successfully handle complex contact dynamics without explicit modeling~\cite{levine2016, kalashnikov2018}. Additional sensing modalities can further improve performance when vision alone is insufficient: Lee et al. combine vision and haptic feedback for peg insertion across variations in geometry and clearance~\cite{lee2020}, while Calandra et al. demonstrate visuo-tactile grasping and re-grasping rather than sustained in-contact assembly~\cite{calandra2018}.

In addition to real-world data acquisition in the previous studies, simulation-based learning has been explored to improve scalability. Techniques such as domain randomization with noise injection aim to bridge the sim-to-real gap by exposing policies to environmental variations during training~\cite{chen2022, laskey2017}. Several works have demonstrated zero-shot sim-to-real transfer using dynamics parameter randomization and sensory noise injection~\cite{andrychowicz2020, peng2018}. However, accurately modeling contact interactions remains challenging, as commonly used rigid-body contact models can exhibit inaccuracies and fail to capture the complexity of real-world interactions~\cite{fazeli2019, peng2018}.

More recently, generative modeling for policy learning, particularly diffusion policies, has been proposed to model multimodal action distributions and improve training stability in complex manipulation tasks learned from demonstrations~\cite{chi2024a, chi2024b}. These methods are particularly well-suited for contact-rich manipulation, where multiple valid contact instances may exist for the same task, making unimodal policies insufficient to capture the underlying solution space~\cite{mason2001, zou2025}. Furthermore, recent work has demonstrated that diffusion-based policies can be extended to incorporate compliance behavior, sustaining contact forces in contact-rich manipulation tasks~\cite{hou2025}. Additional studies have applied these policies to various manipulation tasks involving contact and force feedback, such as object sliding, wiping, or bimanual object manipulation~\cite{wang2025, li2024, xue2025}. 

However, many of these tasks involve relatively unconstrained interactions, where moderate deviations in contact conditions can still lead to successful outcomes. In contrast, some contact-rich manipulation tasks, such as insertion, impose stricter geometric and contact constraints, requiring precise alignment and sustained interaction under tight-clearance conditions~\cite{wang2019}. These challenges become even more prominent in construction-scale tasks, where higher forces and material imperfections further complicate reliable task execution.

Many sensory--motor learning methods rely on time-consuming real-world datasets or large-scale self-supervised real-world interaction via trial-and-error~\cite{levine2016, kalashnikov2018, chi2024a}, which is often impractical in construction settings due to labor cost, human safety, and material degradation. Furthermore, sim-to-real transfer remains unreliable under tight tolerances and unmodeled contact dynamics, limiting the direct applicability of purely learning-based solutions in construction-scale tasks~\cite{zhang2023, parmar2021}. Additionally, many learning-based methods typically operate on tabletop robots in structured environments that do not capture the combined scale, contact forces, material variability, and contact dynamic uncertainty of construction assembly.

\subsection{Learning-integrated model-based control}
\label{sec:Learning_model_based_control}

Learning-based methods have been integrated with model-based control to compensate for modeling errors or optimize controller parameters. Beltran et al. combine reinforcement learning with conventional force--position controllers integrated with a fail-safe mechanism for safe deployment on a real rigid robot manipulator~\cite{beltran2020}. Additionally, Khadar et al. develop a stability-guaranteed reinforcement learning approach that embeds variable impedance controllers within the policy parameterization and enforces Lyapunov stability throughout learning and execution, demonstrating successful application of the peg-in-hole case study~\cite{khader2020}. Shaw et al. also combine variable impedance control with Riemannian Motion Policy components for collision and joint-limit avoidance, improving behavior around these constraints without establishing a general safety guarantee for contact-rich tasks~\cite{shaw2022}. Recently, Hou et al. learn spatially and temporally varying approximate compliance from demonstrations and combine it with diffusion-guided motion to maintain contact modes~\cite{hou2025}. Furthermore, integrated learning--control architectures can also improve sim-to-real transfer in contact-rich tasks. For example, Zhang et al. report that direct transfer of a simulation-trained assembly policy using online admittance-residual learning increases the success rates~\cite{zhang2023}.

\subsection{Summary}

The reviewed literature establishes complementary foundations for contact-rich robotic assembly. Research in robotic assembly in construction has demonstrated the feasibility of full-scale component manipulation and has progressively incorporated planning, perception, and learned policies to compensate for geometric deviations and fabrication uncertainty in assembly tasks. However, correcting component poses or updating planned trajectories does not, by itself, provide online action correction based on the interaction forces that arise during tight-clearance mating. Model-based controllers provide mechanisms for compliant interaction and online pose correction. Depending on their formulation, however, their performance can rely on sufficiently accurate task representations and task-specific controller design or tuning. Learning-based policies, such as diffusion policies, can capture complex sensory--motor relationships that are difficult to specify analytically. Nevertheless, collecting physical demonstrations can be costly and time-intensive for construction-scale tasks, while simulation-trained policies can remain sensitive to discrepancies in geometry, friction, compliance, and contact behavior between simulated and physical systems. Recent research on learning-integrated model-based control demonstrates that learned robot actions can be combined with compliant or admittance-based contact response to improve performance.

\section{Methods}\label{sec:methods}

This section presents our framework for enabling a zero-shot sim-to-real policy transfer for force-aware contact-rich manipulation in construction. We first describe the experimental setup in real-world and simulation environments, including hardware and software (Section~\ref{sec:setup}). We then introduce a pipeline for generating motion-planned trajectories as demonstration datasets in simulation (Section~\ref{sec:data_collection}), which are then used to train diffusion-based policies as our baseline method (Section~\ref{sec:policy_learning}). Finally, we design an adaptive controller (Section~\ref{sec:controller}) that corrects predicted actions from the baseline policies by incorporating real-time force and position feedback to improve execution performance, implement force-aware action correction, and enable zero-shot sim-to-real transfer.

\begin{figure}[!h]
    \centering
    \includegraphics[width=1\linewidth]{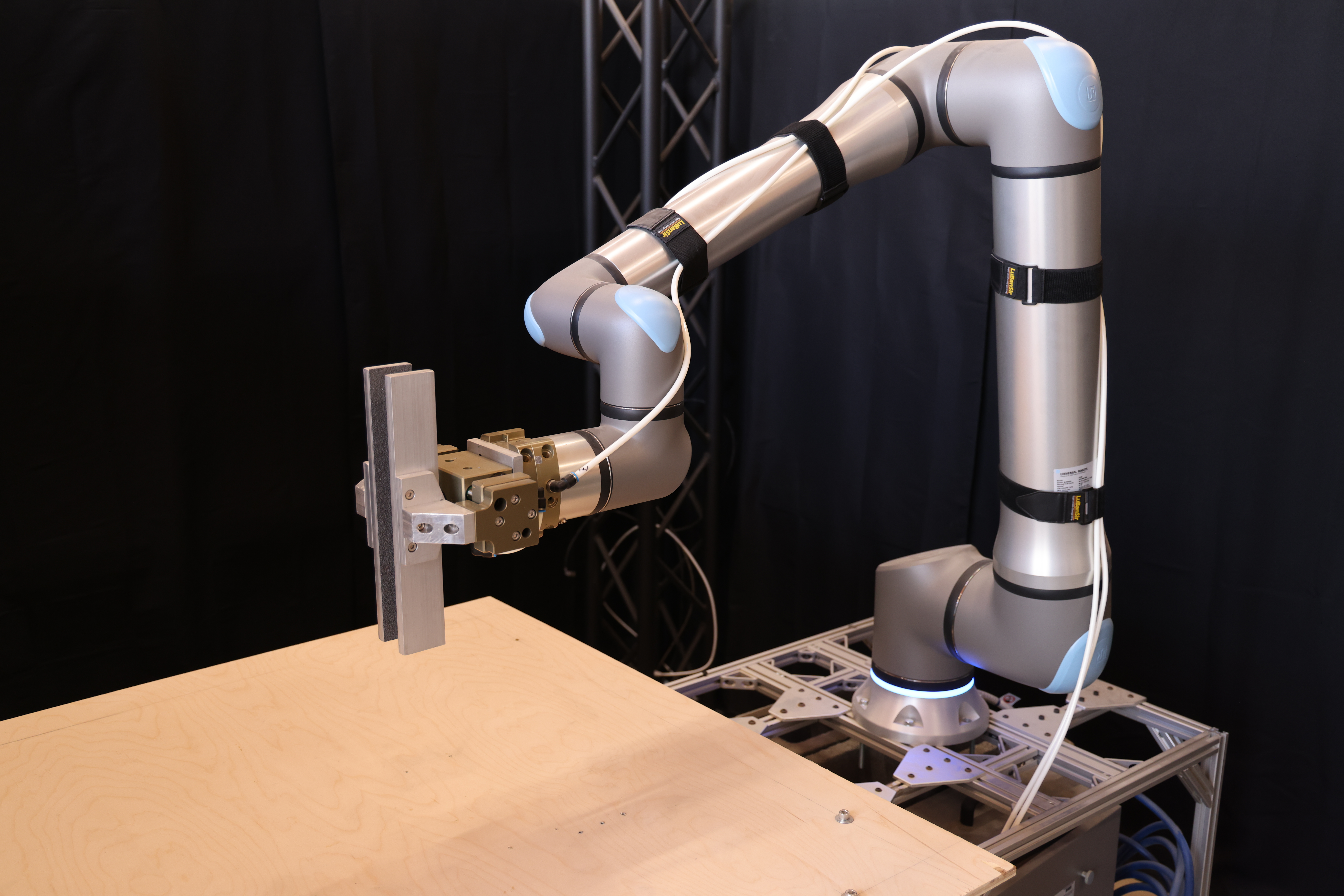}
    \caption{The six-axis UR20 arm and custom end effector.}
    \label{fig:ur_gripper}
\end{figure}

\subsection{Experimental setup}\label{sec:setup}

Our experimental platform comprises a six-axis UR20 robotic arm\footnote{Universal Robot UR20 \cite{ur_20}} equipped with a built-in flange-mounted six-axis F/T sensor and a custom pneumatically actuated parallel gripper \footnote{Schunk PSH-32-1 Parallel Gripper~\cite{schunk_gripper}} (\cref{fig:ur_gripper}). The UR20 arm provides a reach of 1750~\si{\milli\meter} and a rated payload capacity of 20~\si{\kilogram}, with operation up to 25~\si{\kilogram} permitted under the boundary conditions specified by the manufacturer. In the experiments, the combined mass of the end effector and manipulated components was between 6 and 15.2~\si{\kilogram}. The gripper was operated at a fixed pneumatic pressure to maintain a stable and repeatable grasp throughout each trial, without online modulation of the gripping force.

For simulations, we use NVIDIA Isaac Sim 5.0~\cite{isaacsim}. The GPU-accelerated PhysX backend of Isaac Sim enables multiple parallel simulations of contact-rich robotic manipulation, and its native support for Unified Robot Description Format (URDF)-based robot models and six-axis F/T sensing allows the simulated platform to construct a digital representation of the real hardware and the environment without custom physics or sensor modeling. The simulator runs on a dedicated workstation\footnote{Dell Alienware 16 Area-51 Gaming Laptop \cite{dell_laptop}} equipped with an NVIDIA RTX 5090 GPU, an Intel Core Ultra 9 275HX processor, and 32 GB of RAM, enabling parallel simulation of multiple physics-based environments. The UR20 manipulator is modeled in Isaac Sim using its original kinematic and dynamic parameters extracted from the URDF provided by Universal Robots~\cite{ur_20}. The custom gripper is modeled to match the real hardware, and a simulated six-axis rigid-body F/T sensor is integrated at the wrist flange to replicate the force sensing configuration of the physical UR20. The components used for each task are modeled with nominal geometric dimensions and closely matched material properties as their real-world counterparts (see~\Cref{sec:single_experiments} and~\Cref{sec:full_description} for details). 

The simulation therefore reproduces the nominal task geometry and the observation and action interfaces used by the physical system, but it does not assume exact reproduction of fabrication deviations, compliance, friction, or other contact dynamics. These discrepancies constitute the sim-to-real mismatch addressed by the proposed adaptive controller. After training in simulation, the diffusion policy is deployed on the physical platform without task-specific fine-tuning. The online action corrections generated by the adaptive controller remained active during deployment and are part of the proposed method.

\subsection{Data collection}\label{sec:data_collection}

We study contact-rich manipulation tasks in construction (details in~\Cref{sec:experiments}) that require precise geometric alignment under contact forces and robustness to small deviations during real-world execution. Each task is formulated as an assembly problem in which the robot brings a grasped component near its target pre-insertion location, then uses controlled actions to complete the contact-rich insertion. The contact-rich phase is the most challenging, where small pose errors can lead to misalignment of mating parts, robot jamming, or task failure. Therefore, the process of dataset generation in simulation focuses on producing feasible motions from varied pre-insertion configurations to a fully assembled state, so that the policy is exposed to contact and trajectory variations \cite{mandlekar}.

We generate motion-planned demonstration datasets in the simulation environment of NVIDIA Isaac Sim 5.0~\cite{isaacsim}. For each task, the robot arm moves a grasped male component into a horizontal female component rigidly mounted to a tabletop. For each demonstration, the robot executes a two-phase motion: an~\emph{approach phase} that moves the vertical component from a randomized start position to a position directly above the goal, followed by an~\emph{insertion phase} that drives the male into the female component. During each demonstration, we record the timestamped robot position at the flange center point and wrench (i.e., six-axis F/T data) at the flange. The end effector orientation and goal position are kept unchanged during demonstrations to isolate the effect of the contact dynamics during the high-precision insertions.

\begin{figure*}[!h]
    \centering
    \includegraphics[width=0.8\linewidth]{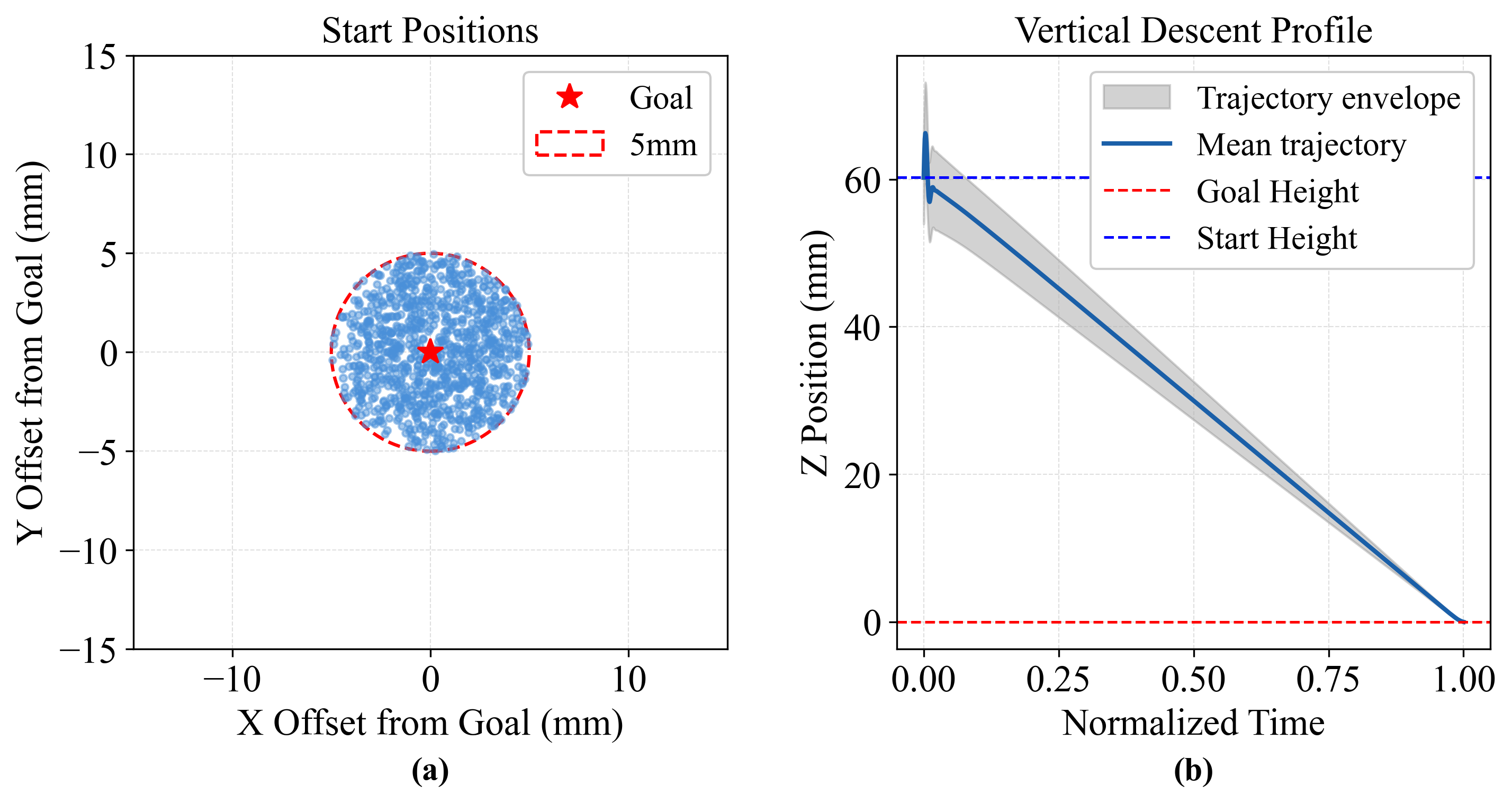} 
    \caption{Example of sampling for trajectory generation (a) horizontal starting offset in the $xy-$plane with $r_{samp}=$ 5 mm, and (b) vertical starting offset and descending heights during the insertion phase with $z_{app}=$ 60 mm and $z_{max}=$ 20 mm.}
    \label{fig:dataset}
\end{figure*}

We randomize the initial position of the gripped element at the beginning of the approach phase. This position is initialized by uniformly sampling the $xy$ coordinates within a disc of radius $r_{samp}$ centered at the goal position (\cref{fig:dataset} (a)), while the vertical coordinate is perturbed around the nominal approach heights $z_{app}$. Specifically, the initial height is sampled as $z = z_{app} + \delta z$ with $\delta z \sim \mathcal{U}(-z_{\max}, z_{\max})$ (\cref{fig:dataset} (b)). The values (details in~\Cref{sec:single_implementation} and~\Cref{sec:full_implementation}) are selected to induce sufficient variability in both trajectories and resultant contact forces. The variation in the initial position expands coverage of the state distribution during training and has been shown to improve the robustness of sim-to-real transfer~\cite{chen2022}.    

The nominal trajectories in both phases are synthesized at a control rate of 100~\si{\hertz} and are perturbed by injecting low-frequency, band-limited noise. The noise is applied only to the lateral $xy$ coordinates, with a maximum amplitude of 2~\si{\milli\meter}. These values are chosen to preserve high-frequency contact information in the recorded data while ensuring the injected perturbations are large enough to induce contact between mating components. Furthermore, the perturbations generate smooth and physically plausible trajectories that resemble corrective motions rather than perfectly straight-line insertions, thereby increasing demonstration diversity and improving policy robustness to distribution shifts during training~\cite{kasaei2025, zhu2025a, laskey2017}. The corresponding wrench measurements are recorded at 100~\si{\hertz} with timestamps synchronized to the position measurements.

\begin{figure*}[!h]
    \centering
    \includegraphics[width=0.8\linewidth]{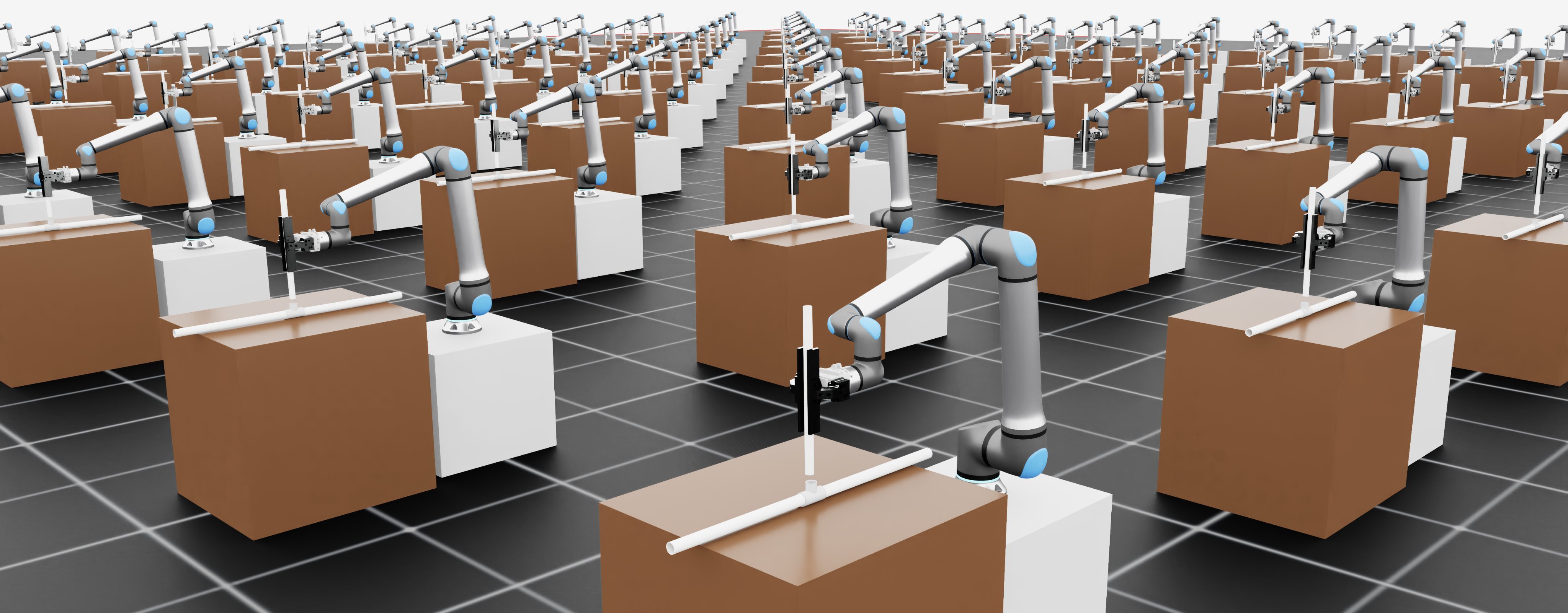}
    \caption{Example of utilizing parallel simulation environments for large-scale data collection.}
    \label{fig:sim_multi_env}
\end{figure*}

The Cartesian trajectory with fixed rotations is tracked using a differential inverse kinematics controller~\cite{ik_controller, heins2021} that outputs joint position targets at each control step. We run parallel simulation environments to accelerate data collection, each with its own trajectory and an independently randomized initial pose~(\Cref{fig:sim_multi_env}). A constant gripper-closed command is applied throughout the demonstrations to maintain a rigid grasp.  

After collecting demonstrations, each trajectory is labeled as successful or failed based on the measured contact forces and the final robot position. A demonstration is marked as successful when two criteria are satisfied: (1) the insertion is fully completed, i.e., the vertical component reaches a predefined goal position within a Euclidean distance of 2~\si{\milli\meter}, and (2) the norm of the measured contact forces remains below the threshold of 1000~\si{\newton} throughout execution. The 2~\si{\milli\meter} positional tolerance accounts for small variations around the fully completed insertion, and the force threshold is intentionally set above the forces typically observed during insertion to retain successful demonstrations that experience higher contact forces, thereby capturing a broader range of contact interactions. Only successful demonstrations are kept for policy training.

All robot positions are transformed to the local coordinate frame of each environment, rendering the dataset invariant to the spatial configuration of parallel simulation environments. The timestamped robot position and wrench are then downsampled to 10~\si{\hertz} to reduce redundancy between consecutive samples and the computational cost of policy training. The demonstration dataset is written to a Hierarchical Data Format version 5 (HDF5) file~\cite{hdf}. This procedure yields a curated dataset of successful, motion-planned trajectories with corresponding force measurements, which is subsequently used to train the diffusion policy described in the following section. 

\subsection{Policy learning}\label{sec:policy_learning}

We train a convolutional neural network (CNN)-based diffusion policy, using a convolutional backbone for the noise prediction network~\cite{chi2024a}. The trained policies serve as the baseline for evaluation and are compared against the proposed method integrating the baseline with the adaptive controller developed in this work to improve robustness during contact-rich interactions, detailed in ~\Cref{sec:controller}.

The diffusion policy is trained on the robot position at the flange center point, $\bm{p} \in \mathbb{R}^3$, and the measured wrench at the flange, $\bm{\mathcal{F}} \in \mathbb{R}^6$. Accordingly, the observation vector at a discrete time step $k$ is defined as:

\begin{equation}
\bm{o}(k)
=
\begin{bmatrix}
\bm{p}(k), &
\bm{\mathcal{F}}(k)
\end{bmatrix}^\top
\in \mathbb{R}^{9}
\end{equation}

At a discrete timestep $k-1$, the policy $\pi$ predicts a sequence of future robot actions $\bm{u}_{pol}(k:k+T_p-1)$ with prediction horizon $T_p$, conditional on a history of observations $\bm{o}(k-T_o+1:k)$ with horizon $T_o$:

\begin{equation}
\bm{u}_{pol}(k:k+T_p-1) = \pi\!\left(\bm{o}(k-T_o+1:k)\right)
\end{equation}

The predicted robot actions $\bm{u}_{pol}(k)  = (\Delta x, \Delta y, \Delta z) \in \mathbb{R}^{3}$ are defined as relative future positions to the latest observed position (referred to as \textit{delta action} in~\cite{chi2024b}). During policy evaluation, only the first $T_a$ steps of $T_p$ are executed to increase the re-planning frequency over the receding horizon. 

The details for policy training, evaluation criteria, as well as their associated parameter values, such as noising/denoising steps $K$, learning rate $L_r$, weight decay $W_d$, batch size $B$, number of epochs $E$, $T_o$, $T_p$, and $T_a$, are presented in~\Cref{sec:single_implementation}. 

\subsection{Adaptive controller}\label{sec:controller}

\begin{figure*}[!h]
    \centering
    \includegraphics[width=1\linewidth]{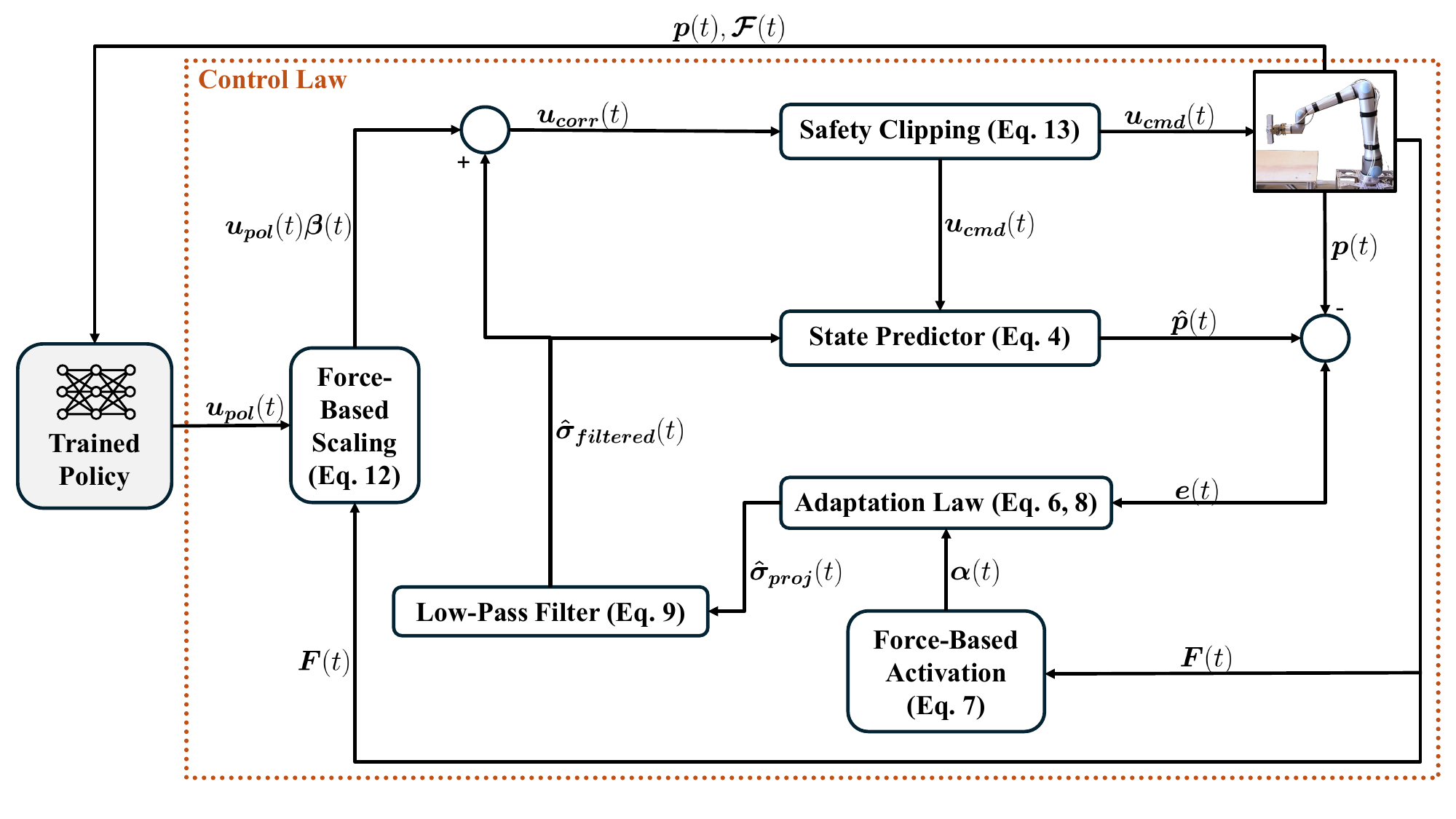}
    \caption{Adaptive controller architecture, including state predictor (Section~\ref{sec:state_predictor}), adaptation law (Section~\ref{sec:adaptation_law}), and control law (Section~\ref{sec:control_law}).}
    \label{fig:controller}
\end{figure*}

At each inference step, the diffusion policy predicts a sequence of actions representing the relative robot positions (see~\Cref{sec:policy_learning}). These actions are corrected by the proposed adaptive controller (shown in~\cref{fig:controller}), which is based on $\mathcal{L}_1$ adaptive control theory introduced in~\cite{hovakimyan2010}. $\mathcal{L}_1$ adaptive control has traditionally been developed and applied in safety-critical dynamical systems, particularly in aerospace applications, where fast adaptation is required to model uncertainty, disturbances, and changing operating conditions~\cite{leman}. These capabilities motivate its application to contact-rich robotic assembly, where the robot encounters uncertain interaction forces, friction, material imperfections, and unmodeled contact dynamics during execution. In our proposed framework, the objective for the controller is to improve robustness during contact by compensating for disturbances on the physical robot, thereby facilitating zero-shot sim-to-real transfer. 

We consider the following dynamic model in continuous time $t$: 

\begin{equation}
        \dot{\bm{p}}(t) = \bm{u}(t) + \bm{\sigma}(t)
\end{equation}

\noindent where $\bm{p}(t) \in \mathbb{R}^3$, $\bm{u}(t) \in \mathbb{R}^3$, and $\bm{\sigma}(t) \in \mathbb{R}^3$ are the system state (robot position at the flange center), control input (policy-predicted action), and unknown disturbances, respectively. This first-order model captures the relationship between the commanded velocity and actual motion, where $\bm{\sigma}(t)$ represents an unknown motion disturbance that captures discrepancies between commanded and realized end-effector motion. In contact-rich scenarios, these discrepancies arise from unmodeled contact forces, material compliance, and geometric misalignment. The controller is aimed at tracking an action generated by the diffusion policy while rejecting disturbances $\bm{\sigma}(t)$. The controller architecture (\cref{fig:controller}) consists of three main components: (1) a state predictor that predicts the system state using a reference model, (2) an adaptation law that estimates disturbances from prediction errors, and (3) a control law that generates corrected actions using a low-pass filter. These components will be discussed in the following sections.

\subsubsection{State predictor}\label{sec:state_predictor}

The state predictor estimates the system state according to: 

\begin{equation}
    \dot{\hat{\bm{p}}}(t) = \bm{A}_{pred} (\hat{\bm{p}}(t) - \bm{p}(t)) + \bm{B}_{pred} \bm{u}_{cmd}(t) + \hat{\bm{\sigma}}_{filtered}(t)
    \label{eq:predictor}
\end{equation}

\noindent where $\hat{\bm{p}}(t) \in \mathbb{R}^3$ is the predicted state, $\bm{u}_{cmd}(t)\in \mathbb{R}^3$ is the control command to the robot, and $ \hat{\bm{\sigma}}_{filtered}(t) \in \mathbb{R}^3$ is the estimated disturbance. The predictor incorporates the control command and disturbance estimate while regulating the prediction error through the predictor bandwidth $\bm{A}_{pred} = a_{pred}\bm{I}$, where $\bm{I} \in \mathbb{R}^3$ is the identity matrix and $a_{pred}<0$. $\bm{B}_{pred} = \bm{I}$ is a unit input gain as the control input directly drives the state rate. The first term in Eq.~(\ref{eq:predictor}) drives $\hat{\bm{p}}(t)$ toward the actual position $\bm{p}(t)$, the second accounts for the commanded motion (discussed in~\Cref{sec:adaptation_law}), and the third injects the filtered estimated disturbance (also discussed in~\Cref{sec:adaptation_law}). This formulation enables the predictor to separate commanded motion from disturbance-induced deviations, which is crucial for isolating the corrections applied by the adaptive controller. We also impose a safety bound $\|\hat{\bm{p}}(t) - \bm{p}(t)\| \leq$ 50~\si{\milli\meter}, serving as a sanity check, if the predictor drifts more than 50~\si{\milli\meter} from the real position, we reset it to $\bm{p}(t)$ to prevent the adaptation from operating on unreliable predictions. 

\subsubsection{Adaptation law}\label{sec:adaptation_law}

The prediction error is defined as: 
\begin{equation}
    \bm{e}(t) = \hat{\bm{p}}(t) - \bm{p}(t)
\end{equation}

This error quantifies the deviation between the predicted and measured states and serves as the feedback signal for disturbance estimation. The magnitude of the prediction error is positively correlated with the level of the unmodeled disturbances. For example, when $\hat{\bm{\sigma}}(t) \approx {\bm\sigma}(t)$, the predictor dynamics match the true system dynamics and the prediction error $\bm{e}(t) \approx \bm{0}$. 

The disturbance estimate is updated according to: 

\begin{equation}
    \dot{\hat{\bm{\sigma}}}(t) = \Gamma \bm{e}(t) \alpha(t) - \lambda \hat{\bm{\sigma}}(t)
\end{equation}

\noindent where $\Gamma > 0$ is the adaptation gain and $\lambda > 0$ is a decay rate that prevents unbounded growth of the disturbance estimate. The parameter $\alpha(t) \in [0, 1]$ is a force-based activation function that modulates the adaptation rate based on the force magnitude, defined as follows \cite{ding2018, orr2002}: 

\begin{equation}
    \alpha(t) = \tanh\left(\frac{||\bm{F}(t)||}{F_0}\right)
\end{equation}

\noindent where $\bm{F}(t) \in \mathbb{R}^3$ is the contact force extracted from the measured wrench $\bm{\mathcal{F}}(t)$ at the flange, $||\bm{F}(t)||$ is the magnitude of the measured force, and $F_0$ is the activation threshold. This saturating nonlinearity gates adaptation based on the contact state. Negligible forces yield $\alpha(t) \approx 0$, suppressing adaptation during free-space motion, while significant contact forces drive $\alpha(t) \approx 1$, fully activating disturbance compensation during the contact phases. To ensure boundedness and prevent drift in the adaptive update, the disturbance estimate is projected~\cite{hovakimyan2010}:

\begin{equation}
    \hat{\bm{\sigma}}_{proj}(t) = \begin{cases}
        \hat{\bm{\sigma}}(t) & if $ $||\hat{\bm{\sigma}}(t)|| \leq \sigma_{\max} \\
        \hat{\bm{\sigma}}(t) \frac{\sigma_{\max}}{||\hat{\bm{\sigma}}(t)||} & otherwise
    \end{cases}
\end{equation}

\noindent where $\sigma_{\max}$ is the maximum disturbance. The projected disturbance estimate with initial condition $\hat{\bm{\sigma}}_{filtered}(0) = \bm{0}\in \mathbb{R}^3$ is then filtered using an exponentially-weighted average \cite{hovakimyan2010}:

\begin{equation}
    \hat{\bm{\sigma}}_{filtered}(k+1) = \gamma \hat{\bm{\sigma}}_{proj}(k) + (1 - \gamma) \hat{\bm{\sigma}}_{filtered}(k)
    \label{eq:sig_filtered}
\end{equation}

\noindent where the filter coefficient $\gamma$ is defines as:

\begin{equation}
    \gamma = \frac{\omega_c \Delta t}{1 + \omega_c \Delta t}
\end{equation}

\noindent with $\omega_c = 2\pi f_c$ and $f_c = 10$~\si{\hertz}, which matches the sampling frequency of the training dataset. This choice attenuates high-frequency noise in the disturbance estimate while avoiding excessive phase lag.

\subsubsection{Control law}\label{sec:control_law}

The corrected control action is computed as:

\begin{equation}
    \bm{u}_{corr}(t) = \bm{u}_{pol}(t) \cdot \beta(t) + \hat{\bm{\sigma}}_{filtered}(t)
\end{equation}

\noindent where $\beta(t)$ is a force-based scaling factor, and $\hat{\bm{\sigma}}_{filtered}(t)$ is the filtered disturbance compensation from Eq.~(\ref{eq:sig_filtered}). This formulation allows the controller to attenuate the policy's predicted actions under high contact forces while injecting corrective actions to compensate for disturbances. The scaling factor reduces action magnitude during high forces and is calculated as \cite{kikuuwe2006}:

\begin{equation}
    \beta(t) = \frac{1}{1 + k_d ||\bm{F}(t)||}
\end{equation}

\noindent where $k_d>0$ is the force damping gain. This inverse relationship promotes compliance: as the magnitude of measured contact force $||F(t)||$ increases, the magnitude of the predicted action decreases, reducing excessive force buildup during contact-rich phases.

To prevent large deviations from the nominal command, the corrected action is projected onto a bounded set:
\begin{equation}
\mathbf{u}_{\mathrm{cmd}}(t)
=
\beta(t)\,\mathbf{u}_{\mathrm{nom}}(t)
+
P_{\mathcal{U}}
\!\left(
\mathbf{u}_{\mathrm{corr}}(t)
-
\beta(t)\,\mathbf{u}_{\mathrm{nom}}(t)
\right)
\label{eq:ucmd_projection}
\end{equation}

\noindent where $P_{\mathcal{U}}:\mathbb{R}^3\to\mathcal{U}$ is a projection operator that denotes Euclidean projection onto $\mathcal{U}
=
\left\{
\mathbf{v}\in\mathbb{R}^3 : \|\mathbf{v}\|_\infty \le \delta_{\max}
\right\}$ and  $\bm{u}_{cmd}$ is the commanded action sent to the robot. For example, for a given vector $\mathbf{z}=[z_1,z_2,z_3]^\top$: 

\begin{equation}
P_{\mathcal{U}}(\mathbf{z})
=
\begin{bmatrix}
g(z_1)\\
g(z_2)\\
g(z_3)
\end{bmatrix},
\quad
g(z_i)
=
\begin{cases}
-\delta_{\max}, & z_i<-\delta_{\max}\\
z_i, & -\delta_{\max}\le z_i\le \delta_{\max}\\
\delta_{\max} & z_i>\delta_{\max}
\end{cases}
\label{eq:proj_componentwise}
\end{equation}

\noindent where $i=1,2,3$. It ensures that each component of the projected vector is bounded within $[-\delta_{\text{max}}, \delta_{\text{max}}]$ . All continuous-time expressions are implemented in discrete time using a forward Euler integration scheme with sampling time $\Delta t$. The parameters used for the adaptive controller are summarized in~\Cref{sec:single_implementation}.

\section{Experiments} \label{sec:experiments}

We evaluate our proposed method through a series of experiments conducted in simulation and real-world environments. The experimental studies consist of two categories of case studies: (1) single assembly experiments on two contact-rich tasks with different geometries, clearances, and material properties (\Cref{sec:single_experiments}), and (2) a full-scale assembly experiment involving multi-stage sequential contact-rich tasks (\Cref{sec:full_experiment}). For each case study, we describe the task setup, implementation details, and evaluation metrics used to assess task success.

\subsection{Single assembly experiments} \label{sec:single_experiments}

This section presents two contact-rich single assembly tasks with the objective of evaluating and benchmarking our proposed method in simulation and real-world environments.

\subsubsection{Task Description}\label{sec:single_description}

\begin{figure}[!h]
    \centering
    \includegraphics[width=1.0\linewidth]{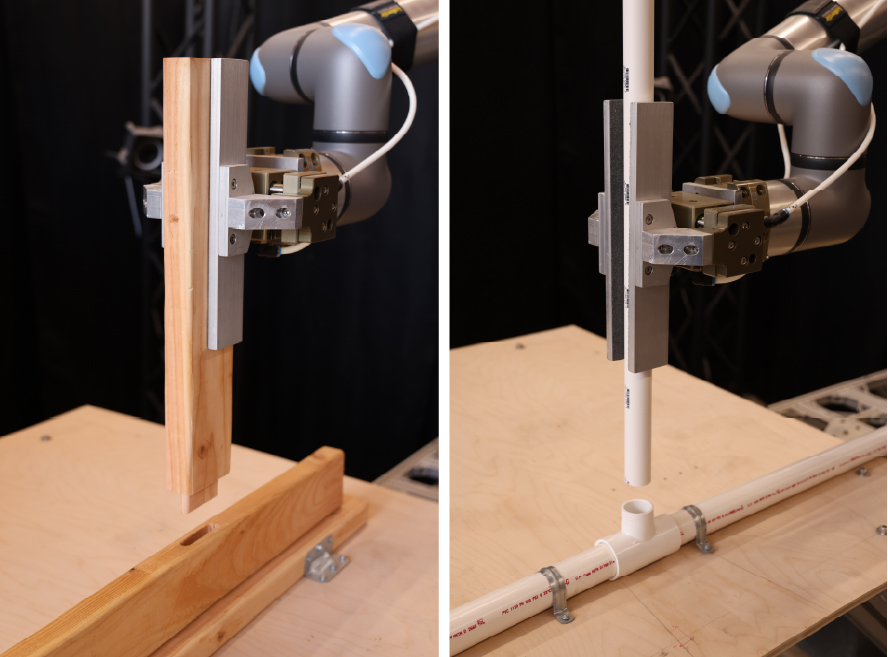}
    \caption{Single contact-rich assembly tasks in construction: mortise and tenon timber joint assembly (left) and pipe fitting (right).}
    \label{fig:single_tasks}
\end{figure}

The single assembly experiments evaluate the proposed framework on two contact-rich manipulation tasks: (I) vertically inserting a timber tenon into its mating mortise, and (II) inserting a PVC pipe segment into one socket of a three-way tee fitting (\cref{fig:single_tasks}). The two tasks differ in geometry, clearance, and material properties, providing a basis to evaluate the proposed framework to generalize across distinct contact-rich assembly scenarios. Prior to each task, the robot is assumed to have already grasped the vertical element and positioned it close to a pre-insertion pose. During data collection, this initial pre-insertion pose is randomized (\Cref{sec:data_collection}). During evaluation, the initial pose is kept constant and centered above the goal position at a vertical distance of 50~\si{\milli\meter}. The end-effector orientation, as well as the goal position, is kept constant during demonstrations and rollouts.

\begin{figure*}[!h]
    \centering
    \includegraphics[width=0.8\linewidth]{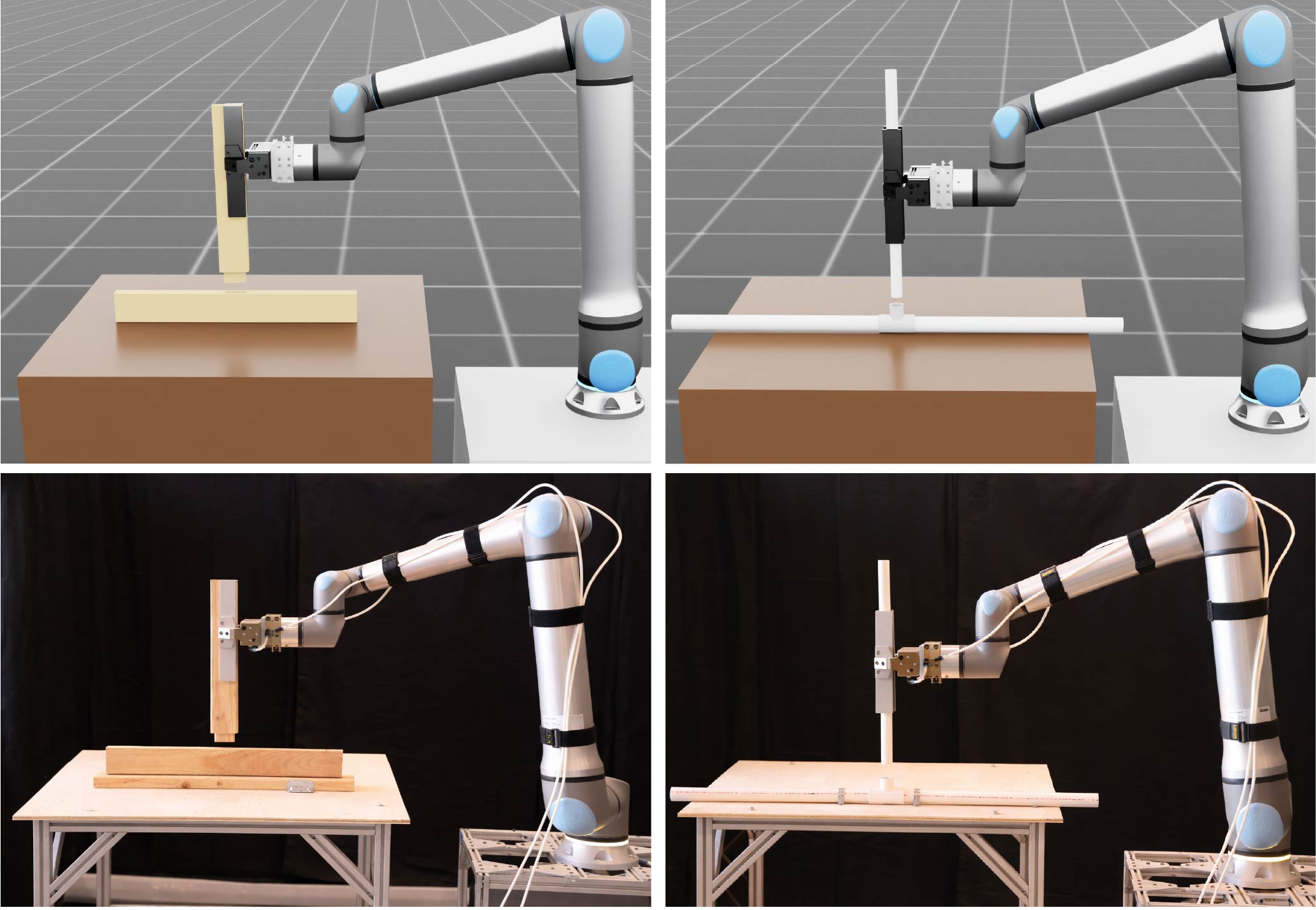}
    \caption{Mortise and tenon assembly in sim and real (left) and pipe fitting in sim and real (right).} 
    \label{fig:sim_real_single}
\end{figure*}

The nominal clearance between the tenon and the mortise is 1~\si{\milli\meter} on each of two opposing sides, whereas the nominal radial clearance between the PVC pipe and the inner wall of the fitting socket is 0.25~\si{\milli\meter}. These clearances are used only during evaluation in simulation and real experiments. The simulated setup reproduces the task-relevant geometry and spatial arrangement of the physical setup: both the manipulated and mating components are modeled as rigid bodies, with the former grasped by the robot end effector and the latter fixed to a stationary table, as shown in \cref{fig:sim_real_single}. To increase the diversity of feasible trajectories in the synthetic demonstration dataset, we use larger clearances during data collection in simulation. For the timber joint, the clearance is maintained at 1~\si{\milli\meter} along the longer side and increased from 1 to 1.2~\si{\milli\meter} along the shorter side; for the pipe-fitting task, the radial clearance is increased from 0.25 to 0.75~\si{\milli\meter}.

\subsubsection{Implementation details}\label{sec:single_implementation}

For each task, we collect 1000 demonstrations in simulation using the data collection procedure described in~\Cref{sec:data_collection} with $r_{samp}=$ 5~\si{\milli\meter}, $z_{app}=$ 60~\si{\milli\meter}, and
$z_{\max} =$ 20~\si{\milli\meter}. The diffusion policies are trained exclusively on simulation data, and four independent models with identical parameters, as summarized in ~\Cref{tab:policy_params}, are trained for each task to account for stochasticity during evaluation.

\begin{table}[!h]
    \centering
    \caption{Parameters used for policy training and execution for both tasks.}
    \label{tab:policy_params}
    \begin{tabular}{llr}
        \toprule
        Symbol & Description & Value \\
        \midrule
        $L_r$ & Learning rate & $2\mathrm{e}{-4}$ \\
        $W_d$ & Weight decay rate & $1\mathrm{e}{-6}$ \\
        $B$ & Batch size & 1024 \\
        $E$ & Number of epochs & 1000 \\
        $T^p_o$ & Observation horizon & 2 \\
        $T_p$ & Action prediction horizon & 16 \\
        $T_a$ & Action execution steps & 3 \\
        $K$ & Noisin/denoising steps & 100 \\
        \bottomrule
    \end{tabular}
\end{table}

We evaluate and compare the following three execution modes in simulation and their zero-shot transfer to the real-world:

\begin{itemize}[itemsep=1pt, topsep=2pt]
\item Baseline: the predicted actions from the trained diffusion policies are executed without correction.
\item Ours: the baseline models are integrated with our proposed adaptive controller, which refines the policy-predicted actions in real time using force and position feedback.
\item Compliant policy: the baseline models are integrated with a compliance controller from~\cite{lee2020} used for benchmarking, which also modifies the policy output during execution. 
\end{itemize}

All controller states are reset at the beginning of each rollout to ensure a fair comparison across all modes.

\Cref{tab:adapt_control_params} and \Cref{tab:comp_control_params} summarize the task-specific parameters used for the adaptive (ours) and compliance (benchmark) controllers, respectively. These parameters are manually tuned in simulation and remain fixed during real-world deployment. Simulated evaluations are carried out using a custom pipeline that enables parallel execution across multiple environments.

\begin{table*}[!h]
    \centering
    \caption{Parameters for our proposed adaptive controller.}
    \label{tab:adapt_control_params}
    \begin{tabular*}{0.7\textwidth}{l@{\extracolsep{\fill}}llrr}
        \toprule
        Symbol & Description & Timber joint & Pipe fitting \\
        \midrule
        $a_{pred}$ & Predictor bandwidth constant & -62.8 & -62.8 \\
        $\Gamma$  & Adaptation gain & 5 & 80 \\
        $\lambda$ & Decay rate & 0.98 & 0.99 \\
        $k_d$ & Force damping gain & 0.005 & 0.005 \\
        $F_0$ & Force activation threshold (N) & 30 & 30 \\
        $\sigma_{\max}$ & Max disturbance (mm) & 1 & 1\\
        $\delta_{\max}$ & Max allowed correction (mm) & 10 & 10\\
        \bottomrule
    \end{tabular*}
\end{table*}

\begin{table}[!h]
    \centering
    \caption{Parameters for the compliance controller adapted from~\cite{lee2020}.}
    \label{tab:comp_control_params}
    \begin{tabular}{llrr}
        \toprule
        Symbol & Description & Timber joint & Pipe fitting \\
        \midrule
        M & Mass (kg) & 700 & 75 \\
        K & Stiffness (kN/m) & 500 & 0.5 \\
        \bottomrule
    \end{tabular}
\end{table}

\subsubsection{Evaluation metrics}\label{sec:single_evaluation}

A rollout is considered successful if the tenon reaches the insertion goal within a Euclidean distance of 3~\si{\milli\meter} (i.e., $\left\| \mathbf{p}(t) - \mathbf{p}_{\mathrm{goal}} \right\| \leq 3~\mathrm{mm}$). Each rollout is executed for at most 600 inference steps and is terminated if either the measured contact force in any Cartesian direction exceeds 200~\si{\newton} (i.e., $\max\left(\left|f_x\right|,\left|f_y\right|,\left|f_z\right|\right) > 200~\mathrm{N}$) or the robot position deviates more than 100~\si{\milli\meter} from the goal position. If the success condition is satisfied before termination, the rollout is marked as successful; otherwise, it is considered a failure.

The performance of each execution mode is reported as the average success rate, computed across rollouts using 4 independently trained policies for each task (see \Cref{sec:single_implementation}). For each trained policy, we execute 50 rollouts in simulation and 10 rollouts on the real robot. As a result, for each execution mode per task we average the success rate for 200 and 40 rollouts in simulation and real-world, respectively. 

In addition to the average success rate, we evaluate forces, $\mathbf{F} \in \mathbb{R}^3$, by analyzing the distribution of maximum contact force magnitudes ($F_{\max} = \max \left\| \mathbf{F} \right\|$)  extracted from the measured F/T data during each rollout. We then perform statistical analysis to compare these distributions across methods. 

Classical one-way ANOVA~\cite{kim2017} and the pooled two-sample Student’s $t$-test~\cite{kim2015}, assume that observations are independent and normally distributed with equal variances. However, the peak force values in real-world experiments exhibit pronounced right skew and heavy tails.(see~\Cref{sec:force_analysis}). We therefore use a rank-based non-parametric test that do not require a Gaussian distribution, but retain assumptions such as independence~\cite{conover1999}.

To assess whether there are statistically significant differences in the peak force distributions for each task across execution modes in simulation and real, we first apply the Kruskal-Wallis test, rank-based non-parametric tests for comparing independent groups~\cite{kruskal1952}. The null hypothesis is that the peak-force observations from the three execution modes follow the same distribution. Given $k = 3$ groups (i.e., execution modes), the test statistic is:

\begin{equation}
    H_1 = \frac{12}{N(N+1)} \sum_{i=1}^{k} n_i{\bar{r}_i^2} - 3(N+1)
\end{equation}

\noindent where $N = \sum_i n_i$ is the total number of observations, $n_i$ is the number of observations in group $i$, and $r_i$ is the average rank of the observations in group $i$. Tied observations are assigned their average rank, and the statistic is corrected for ties as:

\begin{equation}
    C = 1 - \frac{\sum_j \left(t_j^3 - t_j\right)}{N^3 - N},
\end{equation}

\noindent where $t_j$ is the number of observations sharing the same value in the $j$-th set of ties. The corrected test statistic $H_c = \frac{H_1}{C}$ is then used to compute the $p$-value based on its approximate chi-square distribution with $k-1$ degrees of freedom under the null hypothesis.

We then conduct pairwise one-sided Mann-Whitney U tests~\cite{mann1947} regardless of the Kruskal-Wallis test result to determine whether our method produces lower peak forces than the other two execution modes. For each pairwise comparison, we define the alternative hypothesis as:

\begin{equation}
H_1: \theta > 0.5
\end{equation}

\noindent with $\theta$ defined as:


\begin{equation}
\theta =
P(F^{\mathrm{ours}}_{\max} < F^{\mathrm{other}}_{\max})
+ \tfrac{1}{2}
P(F^{\mathrm{ours}}_{\max} = F^{\mathrm{other}}_{\max})
\end{equation}

Here, $\theta$ represents the probability that a randomly selected rollout from our method has a lower peak force than a randomly selected one from the comparison method, with the ties contributing as one-half.  This hypothesis tests whether $\theta$ exceeds 0.5. $\theta = 0.5$ under the null hypothesis of equal distributions. The test statistic $U$ is then defined as:


\begin{equation}
U=\sum_{i=1}^{n_1}\sum_{j=1}^{n_2}
\left[
\mathbf{1}(F^{\mathrm{ours}}_{\max,i}<F^{\mathrm{other}}_{\max,j})
+\tfrac12\mathbf{1}(F^{\mathrm{ours}}_{\max,i}=F^{\mathrm{other}}_{\max,j})
\right]
\end{equation}

\noindent where $\mathbf{1}(\cdot)$ is the indicator function, $n_1$ and $n_2$ are the number of observations from our method and the comparison method respectively. In our case, $n_1 = n_2$ for all pairwise comparisons. To control the family-wise error rate across the two pairwise comparisons (ours vs.\ baseline and ours vs.\ compliant policy), we apply the Bonferroni correction~\cite{dunn1961}, adjusting the significance threshold to $\alpha = 0.05/2 = 0.025$. 

Since the statistical significance alone does not convey the practical magnitude of the differences, we report two complementary effect size measures: (1) The Common Language Effect Size (CLES)~\cite{mcgraw1992} is used as the primary rank-based effect size consistent with the Mann–Whitney U test, while Cohen’s $d$ measure~\cite{cohen1977,goulet2018} is additionally reported as a descriptive measure of the standardized difference in mean peak forces.

The CLES is defined as the probability that a randomly selected force observation from our method is lower than one from the comparison method. Using the relationship to the Mann-Whitney U statistic, this can be expressed as \cite{vargha2000}:

\begin{equation}
    \text{CLES} = \frac{U}{n_1 \cdot n_2}
\end{equation}

\noindent A value of 0.5 indicates no tendency for either method to produce lower forces, while values approaching 1.0 indicate that our method is increasingly likely to produce lower forces.

Cohen's $d$ measures the standardized mean difference as:


\begin{equation}
d = \frac{\mu^{\mathrm{other}}-\mu^{\mathrm{ours}}}{s_p},
\qquad
s_p = \sqrt{\frac{(s^{\mathrm{ours}})^2+(s^{\mathrm{other}})^2}{2}}
\end{equation}

\noindent where $\mu$ and $s$ denote the sample mean and standard deviation, respectively, and $s_p$ is the pooled standard deviation considering $n_1 = n_2$. Positive values of $d$ indicate that our method produces lower mean forces. Because Cohen’s d is based on the sample mean and standard deviation and can therefore be sensitive to skewed and heavy-tailed distributions, we interpret it as a complementary descriptive effect size rather than as the basis for statistical inference.

\subsection{Full-scale assembly experiment}\label{sec:full_experiment}

\begin{figure*}[!ht]
    \centering
    \includegraphics[width=1\linewidth]{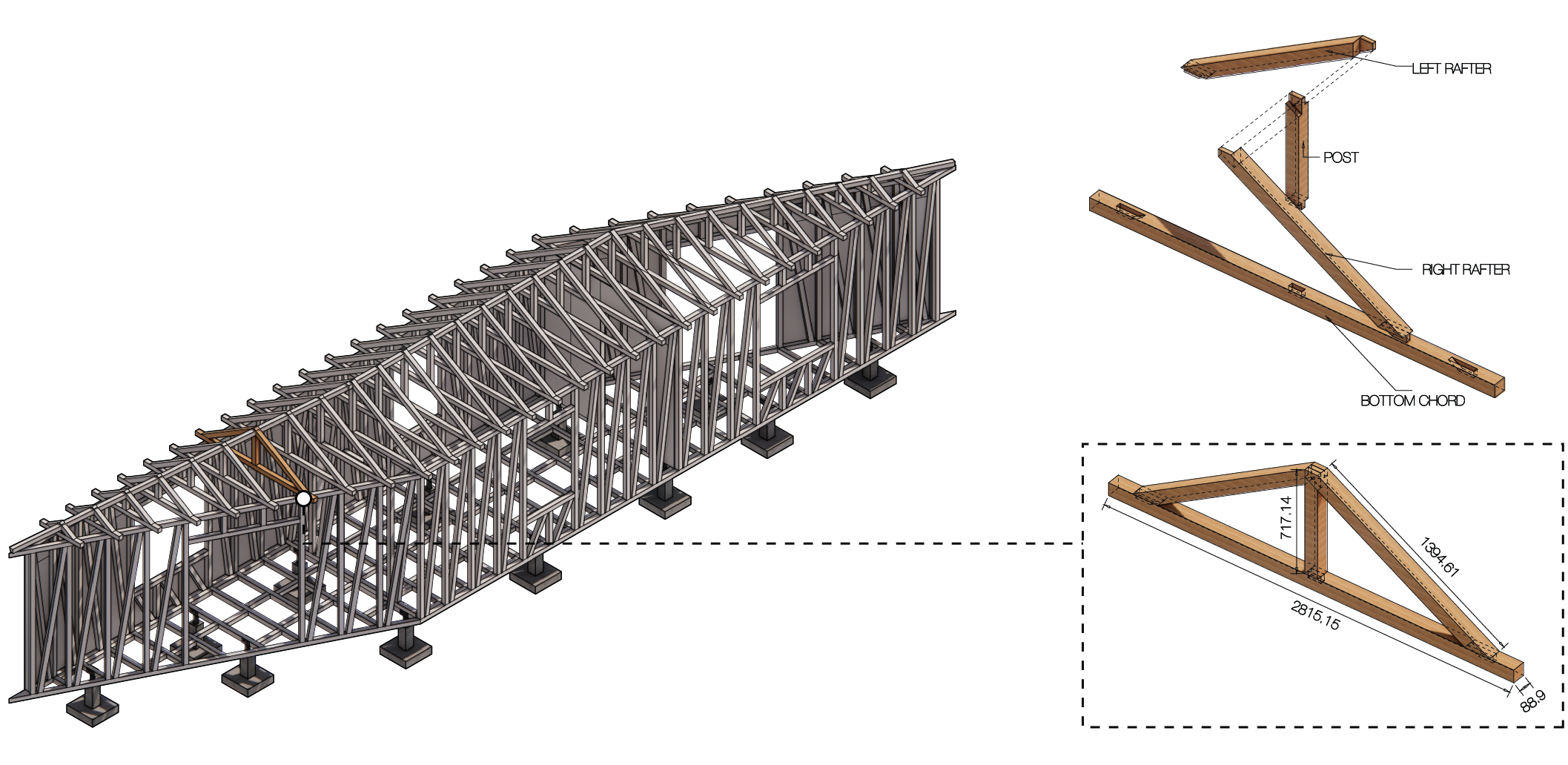}
    \caption{Full-scale assembly of a gable truss: the bottom chord contains one central mortise for the vertical post and two side mortises for the rafters; each rafter is also connected to the post through a lap joint. All dimensions are in millimeters.}
    \label{fig:full_overview}
\end{figure*}

\begin{figure*}[!ht]
    \centering
    \includegraphics[width=1\linewidth]{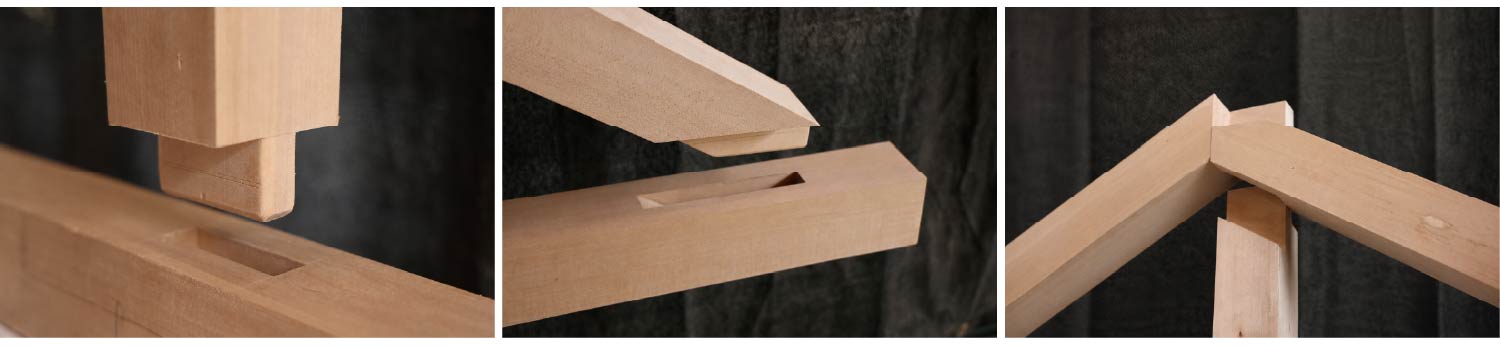}
    \caption{Joint variations in the full-scale assembly of the gable truss: vertical mortise-and-tenon joint between the post and bottom chord (left), angled mortise-and-tenon joint between a rafter and the bottom chord (center), and lap joint between a rafter and the post (right).}
    \label{fig:full_joints}
\end{figure*}

The full-scale assembly experiment utilizes the proposed framework in a multi-step timber construction case study involving three sequential contact-rich assembly tasks with different geometries and contact conditions. The objective of this experiment is to assess the scalability of the proposed framework beyond isolated insertion tasks to construction-scale assembly involving larger components, higher contact forces, and sequential deployment of multiple learned policies.

\subsubsection{Task description}\label{sec:full_description}

We perform the full-scale assembly of a four-member, gable-shaped timber truss as part of a timber-frame structure shown in \cref{fig:full_overview}. The truss comprises a bottom chord, a central post, and two rafters, and incorporates mortise-and-tenon and lap joints~\cite{brown1995, benson1981}. The post and each rafter are inserted into corresponding mortises in the bottom chord, while each rafter is connected to the post near the apex through a lap joint, as shown in \cref{fig:full_joints}. This truss provides a representative building-scale prefabrication task involving large components, tight-fitting joints, and sequential assembly dependencies. It therefore allows us to evaluate whether the proposed framework can extend beyond isolated insertion tasks to the coordinated assembly of a multi-component structural system. Automating such operations could extend digital timber fabrication from component machining to physical assembly, supporting more integrated and repeatable prefabrication workflows while reducing the manual handling, alignment, and fitting of large building components.

\begin{figure*}[h!]
    \centering
    \includegraphics[width=0.8\linewidth]{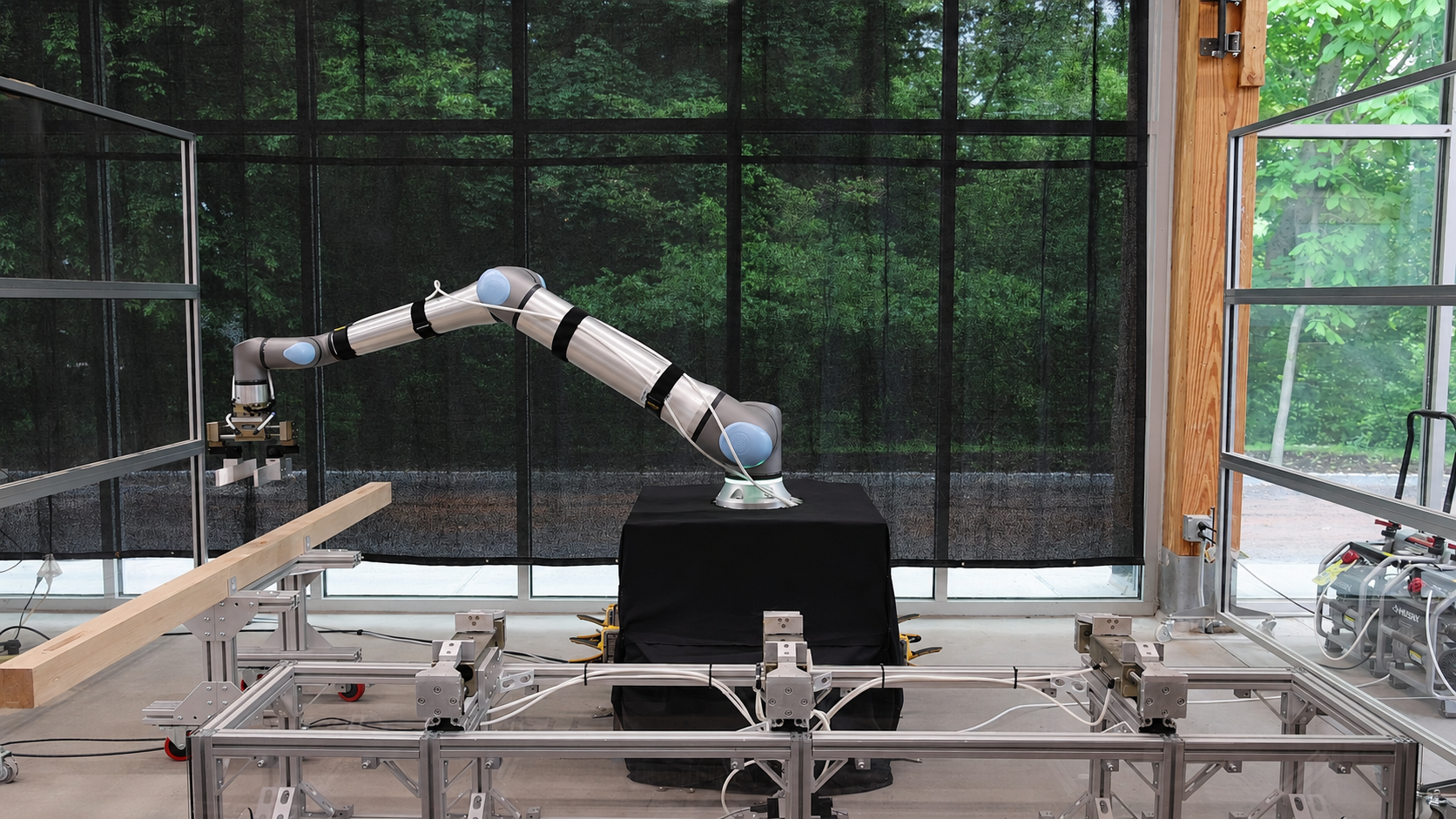}
    \caption{The developed workcell for the full-scale assembly experiment with pick and gripping stations.}
    \label{fig:full_assembly_workcell}
\end{figure*}

\begin{figure*}[ht!]
    \centering
    \includegraphics[width=1\linewidth]{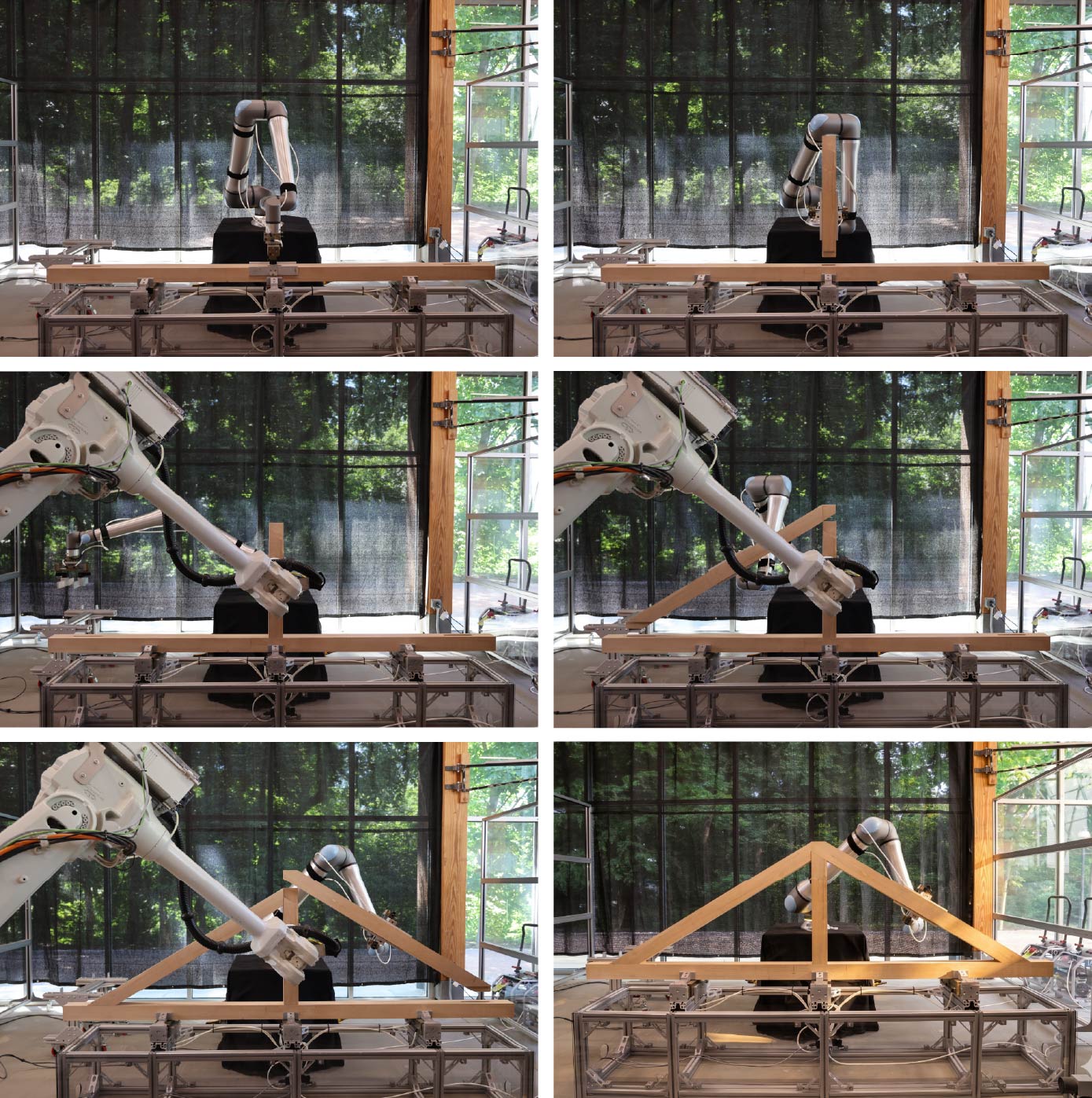}
    \caption{Truss assembly sequence from left to right and top to bottom. First, the bottom chord is placed, followed by the post. Once the post is inserted and released by the UR20, the ABB arm grasps the post to provide support during the subsequent rafter assemblies. The left and right rafters are then assembled sequentially. After the final rafter is inserted, both the post and rafter are released.}
    \label{fig:assembly_sequence}
\end{figure*}

For this full-scale experiment, we developed a workcell (\Cref{fig:full_assembly_workcell}) comprising a pickup station from which the timber elements are grasped and an assembly station equipped with three grippers that secure the bottom chord during assembly. \Cref{fig:assembly_sequence} illustrates the assembly sequence. The bottom chord is first positioned and secured; the UR20 then sequentially assembles the post, left rafter, and right rafter. These insertions constitute the three contact-rich assembly tasks. After the UR20 inserts and releases the post, a second industrial robotic arm\footnote{ABB IRB 4600~\cite{abb}} grasps it to provide stability during the subsequent insertion. Once the right rafter has been inserted, the UR20 releases it, and the supporting robot then releases the post.

In the physical truss, each mortise-and-tenon joint has a nominal clearance of 1~\si{\milli\meter} per side between the tenon and the mortise walls on two opposing sides. As in the single-task experiment, the clearance used in simulation is increased from 1.0 to 1.2~\si{\milli\meter} per side to increase the diversity of feasible trajectories in the synthetic demonstration dataset.

For each of the three contact-rich assembly tasks, the corresponding timber element (either the post or one of the two rafters) is assumed to have already been grasped and positioned at its pre-insertion pose before the robot executes the final insertion motion using the corresponding trained policy integrated with the proposed adaptive controller. The pre-insertion pose is kept constant and centered above the goal position at a vertical distance of 60~\si{\milli\meter} for the post, and 110~\si{\milli\meter} for the rafters.

\subsubsection{Implementation details}\label{sec:full_implementation}

We train three separate diffusion policies for the three contact-rich tasks required for the full-scale assembly experiment described in~\Cref{sec:full_description} using 1000 demonstrations collected per task in simulation, with $r_{samp}=$ 5~\si{\milli\meter}, $z_{app}=$ 110~\si{\milli\meter} for rafters, $z_{app}=$ 60~\si{\milli\meter} for the post, and $z_{\max} =$ 20~\si{\milli\meter}. 

The policy training utilizes the same parameters reported in~\Cref{tab:policy_params}, with a separate task-specific policy trained in simulation for each of the three insertion tasks. Compared with the single-assembly experiments, the larger and heavier timber elements used in the full-scale assembly can produce greater contact forces and can reach the force threshold more easily. Therefore, we adjust the controller parameters to be less reactive to avoid excessive correction during contact while still providing sufficient adaptation for insertion. The controller parameters for this experiment are tuned on the real setup and reported in~\Cref{tab:adapt_control_params_gable}.

\begin{table}[!h]
    \centering
    \caption{Parameters for our proposed adaptive controller for the full-scale gable assembly.}
    \label{tab:adapt_control_params_gable}
    \begin{tabular*}{0.48\textwidth}{l@{\extracolsep{\fill}}llrr}
        \toprule
        Symbol & Description  & Truss assembly\\
        \midrule
        $a_{pred}$ & Predictor bandwidth constant &  -62.8 \\
        $\Gamma$  & Adaptation gain &  1 \\
        $\lambda$ & Decay rate &  0.98 \\
        $k_d$ & Force damping gain &  0.001 \\
        $F_0$ & Force activation threshold (N) &  150\\
        $\sigma_{\max}$ & Max disturbance (mm) &  1\\
        $\delta_{\max}$ & Max allowed correction (mm) & 10\\
        \bottomrule
    \end{tabular*}
\end{table}

\subsubsection{Evaluation metrics}\label{sec:full_evaluation}

The quantitative evaluation of the full-scale assembly focuses on the robustness of the individual contact-rich insertion tasks that constitute the assembly process. Each of the three insertion tasks introduced in~\Cref{sec:full_description}, namely, the insertion of the post, left rafter, and right rafter, is evaluated through 10 independent real-world rollouts using its corresponding policy trained in simulation and the proposed adaptive control framework. The rollout success criteria are consistent with~\Cref{sec:single_evaluation}; however, the force threshold is increased to 300~\si{\newton} to account for the larger and heavier timber elements and higher interaction forces encountered during full-scale assembly. 

The resulting success rates reported in~\Cref{sec:results_full} provide a quantitative assessment of the proposed method for each insertion task, complementing the subsequent evaluation of its integration into the full truss assembly workflow.

\section{Results and discussion}\label{sec:results}

This section presents the evaluation and discussion of results for two sets of case studies: (1) single assembly experiments for timber joint and pipe fitting, and (2) the full-scale truss assembly experiment. 

\begin{table*}[!hb]
\centering
\caption{Average success rates (\%) per execution mode per task in simulation and real-world.}
\label{tab:results}
\begin{tabular*}{0.7\textwidth}{l@{\extracolsep{\fill}}c c c c}
\midrule
Task & \textbf{Ours} & Baseline & Compliant policy & Environment \\
\midrule
Timber joint & \textbf{100} & 87 & 95 & Real \\
Timber joint & \textbf{90} & 73 & 73 & Sim \\
Pipe fitting & \textbf{100} & 100 & 100 & Real \\
Pipe fitting & \textbf{73} & 66 & 69 & Sim \\
\midrule
\end{tabular*}
\end{table*}

\subsection {Single assembly experiments}\label{sec:single_results}

We evaluate our method against the two other execution modes (baseline, and compliant policy) defined in~\Cref{sec:single_implementation}, on the two contact-rich single assembly tasks (mortise and tenon timber joint and pipe fitting) introduced in~\Cref{sec:single_description} in both simulation and real-world settings. Based on the evaluation metrics detailed in~\Cref{sec:single_evaluation}, performance is assessed using the average success rate~(\Cref{sec:avg_sr}) and statistical analysis of peak contact forces~(\Cref{sec:force_analysis}).

\subsubsection{Average success rate} \label{sec:avg_sr}

Table~\ref{tab:results} summarizes the average success rates in simulation and real-world for the two contact-rich manipulation tasks. In simulation, our proposed method consistently outperforms both the baseline and compliant policy for both tasks. 

For mortise and tenon assembly in simulation, the baseline achieves a success rate of 73\%, while the compliant policy does not improve the average success rate. In contrast, our method achieves a higher success rate, indicating improved robustness under tight tolerances. In real-world experiments, the compliant policy achieves 95\% success, which is close to the proposed method (100\%). However, the proposed controller fully eliminates failure cases such as jamming and force-threshold violations observed in the other two methods.

For the pipe fitting task, all execution modes perform similarly, achieving 100\% success in the real-world. This indicates that the task is less sensitive to trajectory misalignment compared to the timber joint. This behavior can be attributed to the material properties of the PVC components, which introduce additional compliance at the contact interface. This inherent compliance allows small misalignments to be accommodated through minor elastic deformation and sliding, effectively relaxing the precision requirements during insertion.

\begin{figure*}[!ht]
    \centering
    \includegraphics[width=1\linewidth]{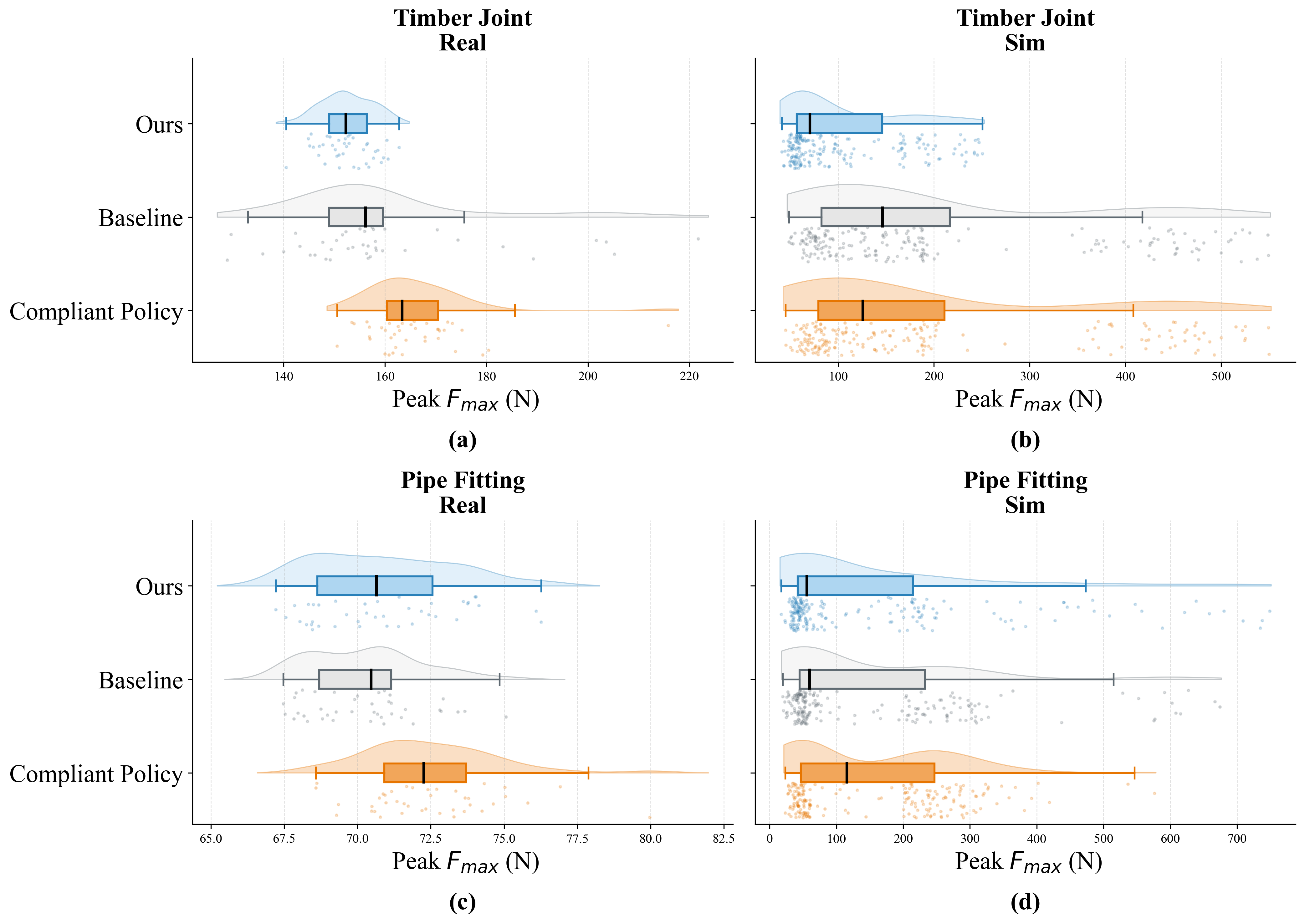} 
    \caption{Peak force distributions across the execution modes for both tasks in simulation and real-world settings.}
    \label{fig:fmag_comparison}
\end{figure*}

\subsubsection{Force analysis}\label{sec:force_analysis}

In addition to the average success rates, we analyze the contact forces for the three execution modes to further highlight the differences between the methods. We present the distributions of peak contact force magnitudes, followed by a statistical analysis across the execution modes. 

~\cref{fig:fmag_comparison} shows the distribution of the maximum contact force magnitudes experienced during the three execution modes in simulation and real-world for both tasks. The Kruskal-Wallis test indicated statistically significant differences among the three execution modes for each task in simulation and real. At 5\% significance level, the $H_C$ statistic was evaluated against a chi-squared distribution with $k-1= 3-1=2$ degrees of freedom (critical value = 5.99). For the timber joint in real, $H=39.47$ with $p<0.001$ and in simulation, $H=83.53$ with $p<0.001$. For the pipe fitting in real, $H=19.00$ with $p<0.001$, and in simulation, $H=6.99$ with $p=0.030$. Since all H values exceeded the critical values and all p-values are below 0.05, the null hypothesis is rejected in all four cases, indicating that at least one execution mode had a different peak force distribution for each task and experimental setting (i.e., real and sim). These results motivated the subsequent pairwise comparisons to identify which execution modes differ most. effects.

~\Cref{tab:stat} summarizes the descriptive statistics and pairwise test results across all experimental conditions. For the timber joint task, the proposed method consistently produces lower peak forces than both the baseline and compliant policy. In the real-world setting, the reduction compared to the compliant policy is statistically significant with a large effect size ($p_{corr}<0.001$, CLES$=0.93$, $d=1.60$), while the difference relative to the Baseline is not significant after correction, likely due to higher variance in the baseline forces. In simulation, both comparisons are highly significant ($p_{corr}<0.001$), with CLES values of 0.74 and 0.72 for the baseline and compliant policy, respectively, indicating a consistent tendency toward lower peak forces.

For the pipe fitting task, differences between methods are less pronounced, consistent with the average success rate results in~\Cref{sec:avg_sr}. In the real-world setting, there is no statistically significant evidence that the proposed method produces lower forces than the baseline ($p_{corr}=1.000$, CLES$=0.43$, $d=-0.30$), but it produces significantly lower forces compared to the compliant policy ($p_{corr}=0.004$, CLES$=0.69$, $d=0.67$). In simulation, statistically significant ($p_{corr}=0.186$), with a negligible standardized mean difference ($d=-0.003$). The compliant-policy comparison is statistically significant ($p_{corr}=0.008$); however, the effect is small (CLES$=0.58$, $d=0.054$), suggesting limited practical improvement.

To summarize, these results show that the benefits of the proposed adaptive controller are most pronounced in tasks involving stiffer timber material, despite its larger geometric clearance. In contrast, for the pipe fitting task with smaller geometric clearance, the lower stiffness and friction of PVC allow misalignments to be accommodated through deformation and sliding, reducing sensitivity to control inaccuracies. This is consistent with the average success-rate results in~\Cref{sec:avg_sr}, where all execution modes achieved similar performance for pipe fitting. As a result, differences between methods are less pronounced. Importantly, even when success rates are similar, force-based analysis reveals meaningful differences in contact quality, highlighting the importance of considering both task performance and contact dynamics when evaluating contact-rich manipulation. Lower contact forces may also reduce the risk of robot wear and material degradation, which is critical in construction applications.

\begin{table*}[!ht]
\centering
\caption{Statistical comparison of peak force magnitude $F_{max}$ (N) across 
methods. $p_{corr}$ is the p-value after Bonferroni correction with significance levels (SL) defined as: *** = $p < 0.001$, ** = $p < 0.01$, * = $p < 0.05$, and ns = not significant.}
\label{tab:stat}
\resizebox{0.9\textwidth}{!}{%
\begin{tabular}{llccccccccc}
\toprule
& & & \multicolumn{2}{c}{$F_{max}$ (N)} & & \multicolumn{5}{c}{One-Sided Mann-Whitney U Test} \\
\cmidrule{4-5} \cmidrule{7-11}
Condition & Method & $N$ & $\mu$ & $s$ && $U$ & $p_{\text{corr}}$ & SL & CLES & Cohen's $d$ \\
\midrule
\multirow{3}{*}{\makecell{Timber joint \\ Real}}
  & Ours             & 40  & 152.5 & 5.0  & & —      & —      & —    & —    & — \\
  & Baseline         & 40  & 159.6 & 20.0 & & 654.0  & 0.161  & ns   & 0.59 & 0.477 \\
  & Compliant policy & 40  & 166.0 & 10.6 & & 107.0  & $<$0.001 & *** & 0.93 & 1.601 \\
\midrule
\multirow{3}{*}{\makecell{Timber joint \\ Sim}}
  & Ours             & 200 & 101.5 & 60.5  & & —      & —      & —    & —    & — \\
  & Baseline         & 200 & 197.1 & 147.0 & & 10485.5 & $<$0.001 & *** & 0.74 & 0.849 \\
  & Compliant policy & 200 & 191.3 & 149.4 & & 11292.0 & $<$0.001 & *** & 0.72 & 0.786 \\
\midrule
\multirow{3}{*}{\makecell{Pipe fitting \\ Real}}
  & Ours             & 40 & 70.9 & 2.4 & & —      & —      & —    & —     & — \\
  & Baseline         & 40 & 70.2 & 1.9 & & 917.0  & 1.000  & ns   & 0.43  & $-$0.304 \\
  & Compliant policy & 40 & 72.5 & 2.2 & & 503.0  & 0.004  & **   & 0.69  & 0.669 \\
\midrule
\multirow{3}{*}{\makecell{Pipe fitting \\ Sim}}
  & Ours             & 200 & 147.1 & 167.5 & & —      & —      & —    & —    & — \\
  & Baseline         & 200 & 146.6 & 149.1 & & 18471.0 & 0.186 & ns  & 0.54 & $-$0.003 \\
  & Compliant policy & 200 & 154.9 & 117.6 & & 16936.5 & 0.008 & **  & 0.58 & 0.054 \\
\bottomrule
\end{tabular}}
\end{table*}

\subsection {Full assembly experiment}\label{sec:results_full}

The results of the full-scale assembly experiment are summarized in~\Cref{tab:full_sr}. The proposed framework achieved 100\% success for the post insertion, 100\% success for the right rafter insertion, and 90\% success for the left rafter insertion across 10 real-world rollouts per task.

In addition to the quantitative evaluation, the full-scale truss assembly described in~\Cref{sec:full_description} was successfully completed using the proposed framework. The learned policies, integrated with the proposed adaptive controller, enabled the sequential insertion of the post and two rafters into the bottom chord, resulting in the assembled gable truss shown in~\Cref{fig:assembly_sequence}. The completed assembly remained stable after the robotic manipulators released the structure.

These results demonstrate that the proposed framework can be extended beyond isolated insertion tasks to a construction-scale assembly workflow involving larger and heavier timber components, higher contact forces, and the sequential deployment of multiple independently trained policies. The successful assembly further highlights the potential of the proposed approach for robotic execution of multi-stage timber construction processes.

\begin{table}[!h]
\centering
\caption{Success rates for 10 rollouts per task in real-world.}
\label{tab:full_sr}
\begin{tabular*}{0.4\textwidth}{l@{\extracolsep{\fill}}c c c c}
\midrule
Joint Name & Success rate (\%) & Environment\\
\midrule
Post & 100 & Real \\
Left Rafter &  90 & Real \\
Right Rafter & 100 & Real \\
\midrule
\end{tabular*}
\end{table}

\section{Conclusion}

This work presents a framework integrating policy learning with an adaptive controller for construction-scale contact-rich robotic assembly. The approach combines large-scale synthetic data generation, diffusion policy learning, and adaptive disturbance compensation to enable zero-shot real-world deployment. By leveraging a simulation pipeline to generate diverse, motion-planned trajectories with force feedback, the framework enables efficient robot learning without relying on costly real-world data collection.

A CNN-based diffusion policy is trained to map multimodal observations to nominal robot actions, while an adaptive control layer refines these actions in real time using position and force feedback. Inspired by $\mathcal{L}_1$ adaptive control, the controller estimates and compensates for residual disturbances arising from unmodeled contact dynamics, friction, and geometric misalignment, enabling robust execution and zero-shot transfer, thereby bridging the sim-to-real gap.

We first evaluate the proposed approach on two representative contact-rich single assembly tasks: mortise and tenon timber joint assembly and pipe fitting. The results demonstrate strong performance across both simulation and real-world settings. In simulation, the adaptive controller consistently improves success rates over both the baseline diffusion policy and a compliance controller, with the largest gains observed in the timber joint task. In real-world experiments, the proposed method achieves perfect success rates across both tasks, demonstrating reliable zero-shot sim-to-real transfer. Furthermore, a full-scale timber truss assembly experiment demonstrates that independently trained policies can be deployed sequentially within a multi-stage assembly workflow involving larger components and higher interaction forces, extending the framework beyond isolated insertion tasks.

Beyond task success, analysis of peak contact forces shows that the proposed method consistently produces lower and more stable interaction forces than both the baseline and compliant policy. While the compliant policy often results in higher peak forces and a wider distribution of contact forces, our method yields more controlled contact behavior, particularly for timber joint assembly. These findings demonstrate that success rate alone does not fully characterize performance in contact-rich manipulation and that force-based metrics provide important insight into contact quality, safety, and robustness. Lower contact forces also reduce the risk of robot wear and material damage, which is critical for reliable deployment in construction applications.

The results further indicate that task performance is influenced by both material properties and geometric tolerances. Stiffer materials, such as timber, are more sensitive to misalignment and therefore benefit substantially from adaptive disturbance compensation. In contrast, although the pipe fitting task has smaller geometric clearances, the lower stiffness and friction of PVC allow misalignments to be accommodated through deformation and sliding, reducing sensitivity to control inaccuracies. As a result, differences between methods are less pronounced. Nevertheless, the proposed method consistently improves contact quality through more stable and controlled interaction dynamics across both tasks.

To summarize, this work demonstrates that integrating diffusion policy learning with an $\mathcal{L}_1$-inspired adaptive controller improves both robustness and execution quality in contact-rich manipulation, enabling reliable zero-shot sim-to-real transfer. The successful deployment of the framework in a full-scale timber truss assembly further demonstrates its scalability to multi-stage construction workflows involving larger components, higher contact forces, and sequential assembly operations. The results of this study highlight a practical pathway toward deploying construction-scale robotic systems in settings where real-world data collection is costly and high-precision execution under uncertainty is critical.

\subsection{Limitations and future work}

While the proposed method demonstrates strong performance in contact-rich assembly tasks, several limitations remain to be addressed in future work. The current framework does not incorporate vision-based perception and assumes a fixed target pose. The framework relies on robot pose and force/torque measurements, which reduces the dimensionality of the input space and enables efficient training. However, this choice may limit generalization to visually complex or occlusion-prone construction environments. Extending the framework to incorporate vision-based perception or richer multimodal representations would enable operation in more unstructured and variable settings.

In addition, this study focuses on representative construction assembly tasks involving mortise-and-tenon timber joinery and pipe fitting, providing a controlled setting to evaluate contact-rich manipulation under tight tolerances. Although these tasks capture key challenges associated with contact-rich assembly, real-world construction scenarios may involve more complex interactions, including multi-axis insertions, compound joint geometries, larger assembly sequences, and coordination across multiple robotic operations. Addressing such scenarios will be important for extending the proposed framework to a broader range of construction applications. In particular, incorporating more expressive learning architectures, such as vision-language-action models (VLAs)~\cite{pi2025, gr2025}, could improve generalization across task variations and enable adaptation to previously unseen assembly conditions and environments.

Moreover, the controller parameters, including adaptation gains and filtering coefficients, are currently tuned manually. While this enables strong performance, it introduces additional engineering effort and limits scalability across tasks and environments. Future work will explore automatic parameter tuning strategies, such as learning-based adaptation, optimization-based tuning, or meta-learning, to enable more generalizable and self-configuring control policies~\cite{beltran2020b, johannsmeier2019}.

Finally, while peak force analysis provides insight into contact quality, additional metrics such as energy consumption, insertion time, and material stress could offer a more comprehensive evaluation of system performance. Expanding the evaluation framework along these dimensions would further strengthen the assessment of contact-rich manipulation systems.

\section*{Acknowledgements}
This research was supported by the Princeton Catalysis Initiative (PCI), the Andlinger Center for Energy and the Environment (Innovative Research in Energy and Environment Convergence Grant), and the School of Architecture at Princeton University. The authors would like to thank members of the Adel Research Group (ARG) for their invaluable contributions to the hardware development and the design of the full-scale experiments.

\section*{Data statement}
The data used in this study are available upon request.

\bibliographystyle{IEEEtran} 
\bibliography{7_bibliography.bib}

\end{document}